\documentclass[10pt,journal,compsoc]{IEEEtran}

\usepackage[nocompress]{cite}

\usepackage{url}
\usepackage{ragged2e}
\usepackage{epsfig}
\usepackage{graphicx}
\usepackage{amsmath,amssymb} 
\usepackage{mathptmx}      
\usepackage{subfigure}
\usepackage{algorithm}
\usepackage{algorithmicx}
\usepackage{algpseudocode}
\usepackage{graphics}
\usepackage{mathrsfs}
\usepackage{threeparttable}
\usepackage{color}
\usepackage[normalem]{ulem}
\usepackage{multirow}
\usepackage{float}
\usepackage{amsfonts}
\usepackage{bm}
\usepackage{array}
\usepackage[table]{xcolor}
\usepackage{colortbl}
\usepackage{pifont}
\usepackage{diagbox}
\usepackage{rotating}
\usepackage{booktabs}
\usepackage{overpic}
\usepackage{textcomp}
\usepackage{contour}
\usepackage{enumitem}
\usepackage{stfloats}
\usepackage{threeparttable}
\usepackage{makecell}
\usepackage[colorlinks=true,breaklinks=true,urlcolor=magenta]{hyperref}

\def\ie{\emph{i.e.}}
\def\eg{\emph{e.g.}}
\def\etc{\emph{etc.}}
\def\etal{{\em et al.~}}

\newcommand{\myPara}[1]{\vspace{6pt}\noindent\textbf{#1}}

\newcommand{\figref}[1]{Fig.~\ref{#1}}
\newcommand{\tabref}[1]{Table~\ref{#1}}

\newcommand{\secref}[1]{\S\ref{#1}}

\newcommand{\tabincell}[2]{\begin{tabular}{@{}#1@{}}#2\end{tabular}}

\definecolor{mygray1}{gray}{.77}
\definecolor{mygray2}{gray}{.92}

\definecolor{myRed}{RGB}{219, 68, 55}
\definecolor{myGreen}{RGB}{15, 157, 88}
\definecolor{myBlue}{RGB}{66, 133, 244}
\newcommand{\tr}[1]{{\textcolor{myRed}{\textbf{#1}}}}
\newcommand{\tg}[1]{{\textcolor{myGreen}{\textbf{#1}}}}
\newcommand{\tb}[1]{{\textcolor{myBlue}{\textbf{#1}}}}
\newcommand{\rev}[1]{{\textcolor{black}{#1}}}

\newcommand{\ourDataset}{CDS2K}

\def\ie{\textit{i.e.}}
\def\eg{\textit{e.g.}}

\graphicspath{{./Imgs/}}
\DeclareGraphicsExtensions{.jpg,.pdf,.png}

\RequirePackage{silence}
\begin{document}

\title{Advances in \\Deep Concealed Scene Understanding}

\author{
  Deng-Ping Fan,~
  Ge-Peng Ji,~
  Peng Xu,~
  Ming-Ming Cheng,~
  Christos Sakaridis,~
  Luc Van Gool\\
\IEEEcompsocitemizethanks{
\IEEEcompsocthanksitem Deng-Ping Fan, Christos Sakaridis, and Luc Van Gool are with the Computer Vision Lab (CVL), ETH Zurich, Zurich, Switzerland.
\IEEEcompsocthanksitem Ge-Peng Ji is with the College of Engineering, Computing \& Cybernetics, ANU, Canberra, Australia.
\IEEEcompsocthanksitem Peng Xu is with the Department of Electronic Engineering, Tsinghua University, Beijing, China.
\IEEEcompsocthanksitem Ming-Ming Cheng is with the Nankai University, Tianjin, China.
}
}


\IEEEtitleabstractindextext{
\begin{abstract} \justifying
Concealed scene understanding (CSU) is a hot computer vision topic 
aiming to perceive objects exhibiting camouflage. 
The current boom in terms of techniques and applications warrants an up-to-date survey. This can help researchers to better understand the global CSU field, 
including both current achievements and remaining challenges.
This paper makes four contributions:
(1) For the first time, we present a comprehensive survey of deep learning techniques aimed at CSU, 
including a taxonomy, task-specific challenges, 
and ongoing developments. 
(2) To allow for an authoritative quantification of the state-of-the-art, 
we offer the largest and latest benchmark for concealed object segmentation (COS). 
(3) To evaluate the generalizability of deep CSU in practical scenarios, 
we collect the largest concealed defect segmentation dataset termed CDS2K 
with the hard cases from diversified industrial scenarios, 
on which we construct a comprehensive benchmark.
(4) We discuss open problems and potential research directions for CSU. 
Our code and datasets are available at \url{https://github.com/DengPingFan/CSU}, 
which will be updated continuously to watch and summarize the advancements 
in this rapidly evolving field.
\end{abstract}

}

\maketitle

\IEEEdisplaynontitleabstractindextext

\IEEEpeerreviewmaketitle

\section{Introduction}\label{sec:introduction}
\IEEEPARstart{C}{oncealed} scene understanding (CSU) aims to recognize objects that exhibit forms of camouflage. By its very nature, CSU clearly is a challenging problem compared with conventional object detection~\cite{fan2023salient,zhao2017pyramid}.
It has numerous real-world applications, including search-and-rescue work, rare species discovery, healthcare (\eg, automatic diagnosis for colorectal polyps~\cite{ji2022vps} and lung lesions~\cite{fan2020infnet}), agriculture (\eg, pest identification~\cite{liu2019pestnet} and fruit ripeness assessment~\cite{rizzo2023fruit}), content creation (\eg, recreational art~\cite{chu2010camouflage}), \etc~
In the past decade,
both academia and industry have widely studied CSU, and various types of images with camouflaged objects have been handled with traditional computer vision and pattern recognition techniques, including hand-engineered patterns (\eg, motion cues~\cite{boult2001into,conte2009algorithm}, optical flow~\cite{hou2011detection,kim2015unsupervised}), heuristic priors (\eg, color~\cite{zhang2016bayesian}, texture~\cite{galun2003texture}, intensity~\cite{tankus1998detection,tankus2001convexity}) and combination techniques~\cite{mittal2004motion,liu2012foreground,li2017foreground}.

\begin{figure}[t!]
  \centering
  \includegraphics[width=.98\linewidth]{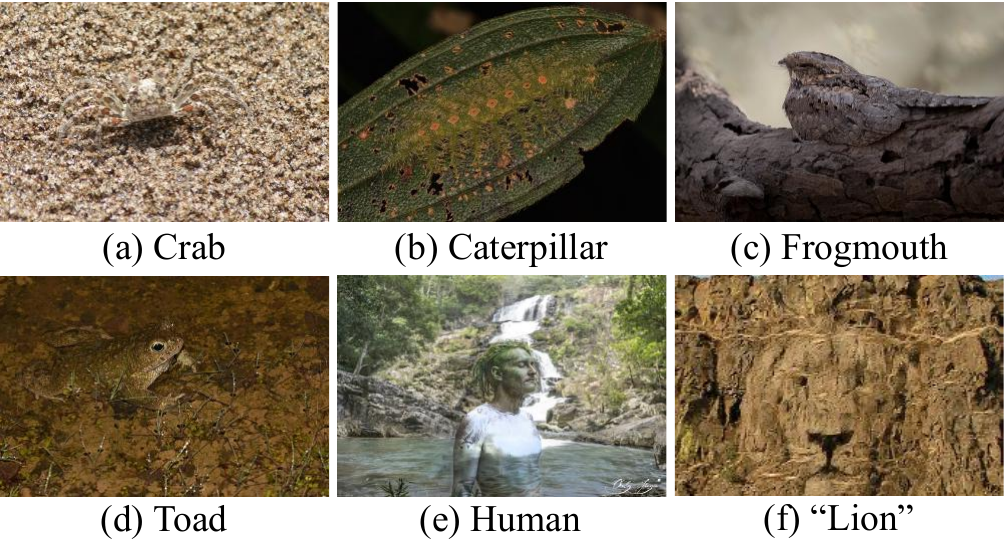}
  \caption{\textbf{Sample gallery of concealed scenarios.} (a-d) show natural animals selected from~\cite{fan2020camouflaged}. (e) depicts a concealed human in art from~\cite{le2019anabranch}. (f) features a synthesized ``lion'' by~\cite{zhang2020deep}.}
  \label{fig:camouflage_sample}
\end{figure}

In recent years, thanks to benchmarks becoming available (\eg, COD10K~\cite{fan2020camouflaged,fan2022concealed} and NC4K~\cite{lv2021simultaneously}) ánd the rapid development of deep learning, this field has made important strides forward.
In 2020, Fan~\etal~\cite{fan2020camouflaged} released the first large-scale public dataset - COD10K - geared towards the advancement of perception tasks having to deal with concealment. This has also inspired other related disciplines. 
For instance, Mei~\etal~\cite{mei2021camouflaged,mei2023distraction} proposed a distraction-aware framework for the segmentation of camouflaged objects, which can be extended to the identification of transparent materials in natural scenes~\cite{yu2022progressive}. In 2023, Ji~\etal~\cite{ji2023gradient} developed an efficient model that learns textures from object-level gradients, and its generalizability has been verified through diverse downstream applications, \eg, medical polyp segmentation and road crack detection.

\rev{Although multiple research teams have addressed tasks concerned with concealed objects, we believe that stronger interactions between the ongoing efforts would be beneficial.}
Thus, we mainly review the state and recent deep learning-based advances of CSU. Meanwhile, we contribute a large-scale concealed defect segmentation dataset termed CDS2K. This dataset consists of hard cases from diverse industrial scenarios, thus providing an effective benchmark for CSU.

{\bf Previous Surveys and Scope.} 
To the best of our knowledge, only a few survey papers were published in the CSU community,
which~\cite{kulchandani2015moving,mondal2020camouflaged}  mainly review non-deep techniques. There are some benchmarks~\cite{bi2022rethinking,caijuan2022survey} with narrow scopes, such as image-level segmentation, where only a few deep methods were evaluated.
In this paper, we present a comprehensive survey of deep learning CSU techniques, thus widening the scope. We also offer more extensive benchmarks with a more comprehensive comparison and with an application-oriented evaluation.

{\bf Contributions.}
Our contributions are summarized as follows:
(1) We represent the initial effort to examine deep learning techniques tailored towards CSU thoroughly. 
This includes an overview of its classification and specific obstacles, as well as an assessment of its advancements during the era of deep learning, achieved through an examination of existing datasets and techniques.
(2) To provide a quantitative evaluation of the current state-of-the-art, we have created a new benchmark for Concealed Object Segmentation (COS), which is a crucial and highly successful area within CSU. This benchmark is the most up-to-date and comprehensive available.
(3) To assess the applicability of deep CSU in real-world scenarios, we have restructured the CDS2K dataset -- the largest dataset for concealed defect segmentation -- to include challenging cases from various industrial settings. We have utilized this updated dataset to create a comprehensive benchmark for evaluation.
(4) Our discussion delves into the present obstacles, available prospects, and future research areas for the CSU community.

\begin{figure*}[t!]
  \centering
  \includegraphics[width=0.9\linewidth]{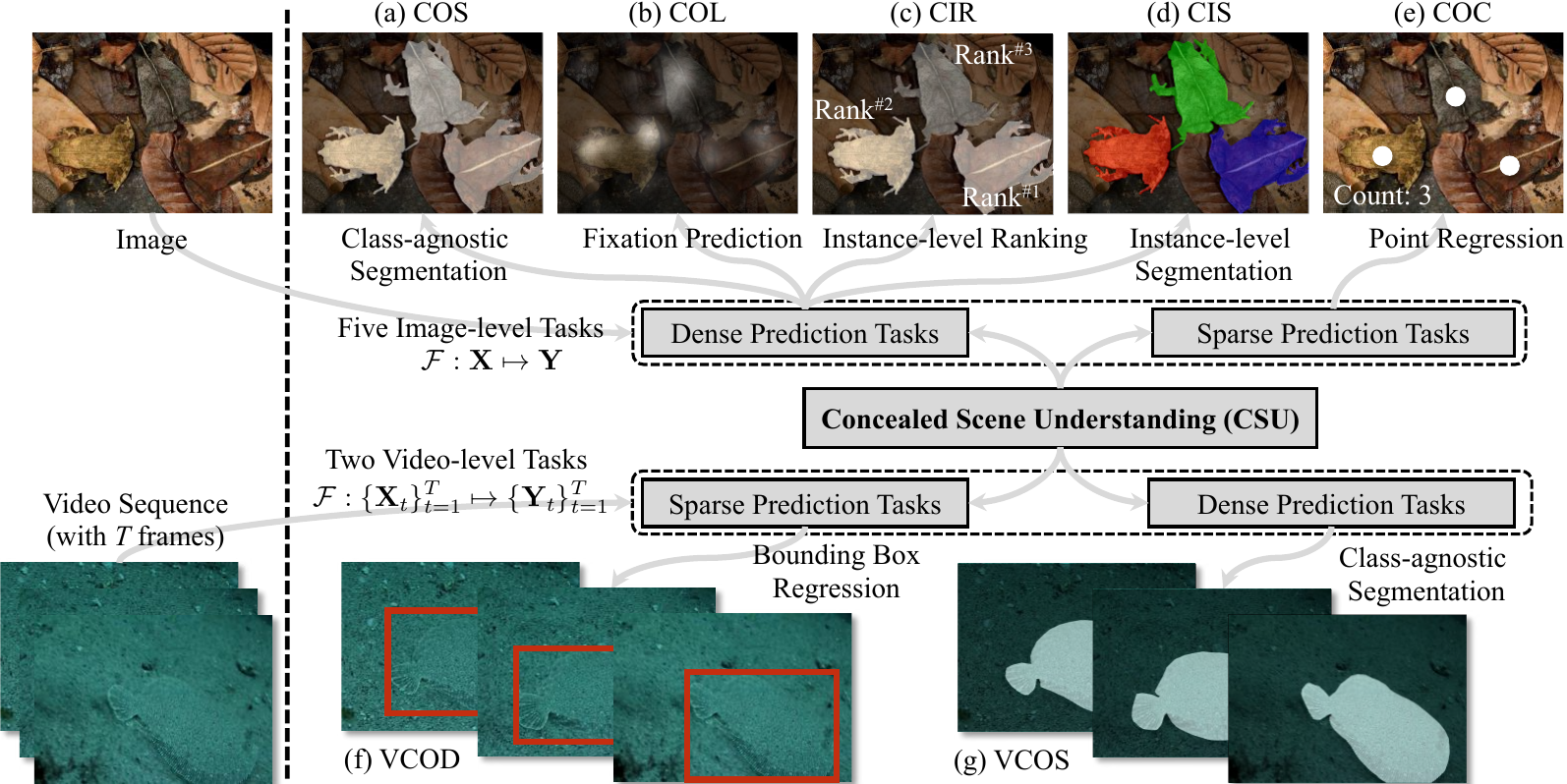}
  \caption{\textbf{Illustration of the representative CSU tasks.} 
  Five of these are image-level tasks: (a) concealed object segmentation (COS), (b) concealed object localization (COL), (c) concealed instance ranking (CIR), (d) concealed instance segmentation (CIS), and (e) concealed object counting (COC). The remaining two are video-level tasks: (f) video concealed object detection (VCOD) and (g) video concealed object segmentation (VCOS). Each task has its own corresponding annotation visualization, which is explained in detail in~\secref{sec:task_taxonomy_and_formulation}.}
  \label{fig:task_definition}
\end{figure*}

\section{Background}\label{sec:background}

\subsection{Task Taxonomy and Formulation}\label{sec:task_taxonomy_and_formulation}

\subsubsection{Image-level CSU}

In this section, we introduce five commonly used image-level CSU tasks, which can be formulated as a mapping function $\mathbf{F}: \mathbf{X} \mapsto \mathbf{Y}$ that converts the input space $\mathbf{X}$ into the target space $\mathbf{Y}$.

\noindent$\bullet$
\textbf{Concealed Object Segmentation (COS)} \cite{fan2022concealed,ji2023gradient} is a class-agnostic dense prediction task, segmenting concealed regions or objects with unknown categories. As presented in~\figref{fig:task_definition} (a), the model $\mathbf{F}_{\text{COS}}\!:~\mathbf{X}~\mapsto~\mathbf{Y}$ is supervised by a binary mask $\mathbf{Y}$ {to predict} a probability $\mathbf{p} \in [0,1]$ for each pixel $\mathbf{x}$ of image $\mathbf{X}$, which is the confidence level that the model determines whether $\mathbf{x}$ belongs to the concealed region.

\myPara{Concealed Object Localization (COL)}~\cite{lv2023towards,lv2021simultaneously} aims to identify the most noticeable region of concealed objects, which is in line with human perception psychology~\cite{lv2023towards}. This task is to learn a dense mapping $\mathbf{F}_{\text{COL}}: \mathbf{X} \mapsto \mathbf{Y}$. The output $\mathbf{Y}$ is a non-binary fixation map captured by an eye tracker device, as illustrated in~\figref{fig:task_definition} (b). Essentially, the probability prediction $\mathbf{p} \in [0,1]$ for 
a pixel $\mathbf{x}$ indicates how conspicuous its camouflage is.

\myPara{Concealed Instance Ranking (CIR)} \cite{lv2023towards,lv2021simultaneously}
is to rank different instances in a concealed scene based on their detectability. The level of camouflage is used as the basis for this ranking. The objective of the CIR task is to learn a dense mapping $\mathbf{F}_{\text{CIR}}: \mathbf{X} \mapsto \mathbf{Y}$ between the input space $\mathbf{X}$ and the camouflage ranking space $\mathbf{Y}$, where $\mathbf{Y}$ represents per-pixel annotations for each instance with corresponding rank levels. 
For example, in~\figref{fig:task_definition}~(c), there are three toads with different camouflage levels, and their  ranking labels are from~\cite{lv2021simultaneously}.
To perform this task, one can replace the category ID for each instance with rank labels in instance segmentation models like Mask R-CNN~\cite{he2017mask}.

\myPara{Concealed Instance Segmentation (CIS)}~\cite{pei2022osformer,le2022camouflaged} is a technique that aims to identify instances in concealed scenarios based on their semantic characteristics. Unlike general instance segmentation~\cite{xie2021polarmask++,chen2020blendmask}, where each instance is assigned a category label, CIS recognizes the attributes of concealed objects to distinguish between different entities more effectively. To achieve this, CIS employs a mapping function $\mathbf{F}_{\text{CIS}}: \mathbf{X} \mapsto \mathbf{Y}$, where $\mathbf{Y}$ is a scalar set comprising various entities used to parse each pixel. This concept is illustrated in~\figref{fig:task_definition} (d).

\myPara{Concealed Object Counting (COC)}~\cite{sun2023ioc}
is a newly emerging topic in CSU that aims to estimate the number of instances concealed within their surroundings. As illustrated in~\figref{fig:task_definition}~(e), the COC is to estimate center coordinates for each instance and generate their counts.  Its formulation can be defined as $\mathbf{F}_{\text{COC}}: \mathbf{X} \mapsto \mathbf{Y}$, where $\mathbf{X}$ is the input image and $\mathbf{Y}$ represents the output density map that indicates the concealed instances in scenes.

Overall, the image-level CSU tasks can be categorized into two groups based on their semantics: object-level (COS and COL) and instance-level (CIR, COC, and CIS). Object-level tasks focus on perceiving objects while instance-level ones aim to recognize semantics to distinguish different entities. Additionally, COC is regarded as a sparse prediction task based on its output form, whereas the others belong to dense prediction tasks. 
Among the literature reviewed in~\tabref{tab:image_csu_work_review}, COS has been extensively researched while research on the other three tasks is gradually increasing.

\subsubsection{Video-level CSU}

Given a video clip $\{\mathbf{X}_{t}\}_{t=1}^{T}$ containing $T$ concealed frames, video-level CSU can be formulated as a mapping function $\mathbf{F}: \{\mathbf{X}_{t}\}_{t=1}^{T} \mapsto \{\mathbf{Y}_{t}\}_{t=1}^{T}$ for parsing dense spatial-temporal correspondences, where $\mathbf{Y}_{t}$ is the label of frame $\mathbf{X}_{t}$. 

\myPara{Video Concealed Object Detection (VCOD)}~\cite{lamdouar2020betrayed}
is similar to video object detection~\cite{jiao2021new}. This task aims to identify and locate concealed objects within a video by learning a spatial-temporal mapping function $\mathbf{F}_{\text{VCOD}}: \{\mathbf{X}_{t}\}_{t=1}^{T} \mapsto \{\mathbf{Y}_{t}\}_{t=1}^{T}$ that predicts the location $\mathbf{Y}_{t}$ of an object  for each frame $\mathbf{X}_{t}$. The location label $\mathbf{Y}_t$ is provided as a bounding box (See \figref{fig:task_definition} (f)) consisting of four numbers $(x,y,w,h)$ indicating the target's location. Here, $(x,y)$ represents its top-left coordinate, while $w$ and $h$ denote its width and height, respectively.

\myPara{Video Concealed Object Segmentation (VCOS)} \cite{cheng2022implicit}
originated from the task of camouflaged object discovery~\cite{lamdouar2020betrayed}. 
Its goal is to segment concealed objects within a video. This task usually utilizes spatial-temporal cues to drive the models to learn the mapping $\mathbf{F}_{\text{VCOS}}: \{\mathbf{X}_{t}\}_{t=1}^{T} \mapsto \{\mathbf{Y}_{t}\}_{t=1}^{T}$ between input frames $\mathbf{X}_{t}$ and corresponding segmentation mask labels $\mathbf{Y}_{t}$. \figref{fig:task_definition} (g) shows an example of its segmentation mask.

In general, compared to image-level CSU, video-level CSU  is developing relatively slowly.
Because collecting and annotating video data is labor-intensive and time-consuming.
However, with the establishment of the first large-scale VCOS benchmark on MoCA-Mask~\cite{cheng2022implicit}, this field has made fundamental progress while still having ample room for exploration.

\subsubsection{Task Relationship}
\rev{Among image-level CSU tasks, the CIR task requires the highest level of understanding as it may not only involve four subtasks, \eg, segmenting pixel-level regions (\ie, COS), counting (\ie, COC), or distinguishing different instances (\ie, CIS), but also ranking these instances according to their fixation probabilities (\ie, COL) under different difficulty levels. Additionally, regarding two video-level tasks, VCOS is a downstream task for VCOD since the segmentation task requires the model to provide pixel-level classification probabilities.}

\subsection{Related Topics}

Next, we will briefly introduce salient object detection (SOD), which, like COS, requires extracting properties of target objects, but one focuses on saliency while the other on the concealed attribute.

\myPara{Image-level SOD} aims to identify the most attractive objects in an image and extract their pixel-accurate silhouettes~\cite{fan2018salient}. 
Various network architectures have been explored in deep SOD models, \eg, multi-layer perceptron~\cite{he2015supercnn,li2015visual,wang2015deep,kim2016shape}, fully convolutional~\cite{zeng2019towards,liu2016dhsnet,wu2019cascaded,zhang2017learning,hou2019deeply}, capsule-based~\cite{zhuge2022salient,liu2019employing,qi2019multi}, transformer-based~\cite{liu2021visual}, and hybrid~\cite{li2016deep,tang2016saliency} networks. 
Meanwhile, different learning strategies are also studied in SOD models, including data-efficient methods (\eg, weakly-supervised with categorical tags~\cite{wang2017learning,li2018weakly,cao2018lateral,li2019supervae,zeng2019multi} and unsupervised with pseudo masks~\cite{zhang2017supervision,zhang2018deep,shin2022unsupervised}), multi-task paradigms (\eg, object subitizing~\cite{he2017delving,islam2018revisiting}, fixation prediction~\cite{wang2018salient,kruthiventi2016saliency}, semantic segmentation~\cite{zeng2019joint,wang2016saliency}, edge detection~\cite{li2018contour,wang2019salient,liu2019simple,zhao2019egnet,su2019selectivity}, image captioning~\cite{zhang2019capsal}), instance-level paradigms~\cite{li2017instance,tian2022learning,fan2019s4net,wu2021regularized}, \etc~To learn more about this field comprehensively, readers can refer to popular surveys or representative studies on visual attention~\cite{borji2012state}, saliency prediction~\cite{borji2019saliency}, co-saliency detection~\cite{fan2022rethinking,fan2020taking,zhang2018review}, RGB SOD~\cite{fan2023salient,borji2019salient,wang2021salient,borji2015salient}, RGB-D (depth) SOD~\cite{zhou2021rgbd,fan2020rethinking}, RGB-T (thermal) SOD~\cite{cong2022does,tu2021multi}, and light-field SOD~\cite{fu2022light}.

\myPara{Video-level SOD.} 
The early development of video salient object detection (VSOD) originated from introducing attention mechanisms in video object segmentation (VOS) tasks. At that stage, the task scenes were relatively simple, containing only one object moving in the video. As moving objects tend to attract visual attention, VOS and VSOD were equivalent tasks. For instance, Wang~\etal~\cite{wang2017video} used a fully convolutional neural network to address the VSOD task. With the development of VOS techniques, researchers introduced more complex scenes (\eg, with complex backgrounds, object movements, and two objects), but the two tasks remained equivalent. 
Thus, later works have exploited semantic-level spatial-temporal features~\cite{le2017deeply,chen2021exploring,le2018video,zhang2021dynamic}, recurrent neural networks~\cite{li2018flow,song2018pyramid}, or offline motion cues such as optical flow~\cite{li2018flow,ji2022fsnet-CVMJ,li2019motion,cong2023psnet}.
However, with the introduction of more challenging video scenes (containing three or more objects, simultaneous camera, and object movements), VOS and VSOD were no longer equivalent. Yet, researchers continued to approach the two tasks as equivalent, ignoring the issue of visual attention allocation in multi-object movement in video scenes, which seriously hindered the development of the field.
To address this issue, in 2019, Fan \etal~introduced eye trackers to mark the changes in visual attention in multi-object movement scenarios, for the first time posing the scientific problem of \textit{attention shift}~\cite{fan2019shifting} in VSOD asks, and constructed the first large-scale VSOD benchmark -- DAVSOD\footnote{\url{https://github.com/DengPingFan/DAVSOD}}, as well as the baseline model SSAV, which propelled VSOD into a new stage of development.

\myPara{Remarks.}
COS and SOD are distinct tasks, but they can mutually benefit via the CamDiff approach~\cite{luo2023camdiff}. This has been demonstrated through adversarial learning~\cite{li2021uncertainty}, leading to joint research efforts such as the recently proposed dichotomous image segmentation~\cite{qin2022highly}. In~\secref{sec:future_research_perspective}, we will explore potential directions for future research in these areas.

\section{Deep CSU Models}\label{sec:approaches}

This section systematically reviews deep CSU approaches based on task definitions and data types. We have also created a GitHub base\footnote{\url{https://github.com/GewelsJI/SINet-V2/blob/main/AWESOME_COD_LIST.md}} as a comprehensive collection to provide the latest information in this field.

\subsection{Image-level CSU Models}\label{sec:image_level_csu_models}

We review the existing four image-level CSU tasks: concealed object segmentation (COS), concealed object localization (COL), concealed instance ranking (CIR), and concealed instance segmentation (CIS). \tabref{tab:image_csu_work_review} summarizes the key features of these reviewed approaches.

\begin{table*}[ht!]
\centering
\caption{
  \textbf{Essential characteristics of reviewed image-based methods.} This summary outlines the key characteristics, including: \textbf{Architecture Design (Arc.)}: The framework used, which can be multi-stream (MSF), bottom-up \& top-down (BTF), or branched (BF) frameworks. \textbf{Multiple Cues (M.C.)}: Whether an auxiliary cue is supplied. \textbf{Supervision Level (S.L.)}: Whether fully-supervised ($\bigstar$) or weakly-supervised ($\diamondsuit$) learning is used. \textbf{Task Level (T.L.)}: The specific tasks addressed by the method, including COS ($\bullet$), CIS ($\circ$), COC ($\square$), and multi-task learning ($\blacksquare$). N/A indicates that the source code is not available. For more detailed descriptions of these characteristics, please refer to~\secref{sec:image_level_csu_models} on image-level CSU models.
}
\label{tab:image_csu_work_review}
\scriptsize
\renewcommand{\arraystretch}{1}
\renewcommand{\tabcolsep}{0.383cm}
\begin{threeparttable}
\begin{tabular}{|  c | r || r | c | cccc | c |} \hline
    \rowcolor{mygray1}
    \textbf{\#} &\textbf{Model}~~ &\textbf{Pub.}
    &\textbf{Core Component} &\textbf{Arc.} &\textbf{M.C.} &\textbf{S.L.} &\textbf{T.L.} &\textbf{Code} \\
    \hline
    \hline
    1 &ANet~\cite{le2019anabranch} &CVIU$_{19}$ 
    &classification \& segmentation streams &BF &\checkmark &$\bigstar$ &$\bullet$ &\href{https://sites.google.com/view/ltnghia/research/camo}{Link}\\ \rowcolor{mygray2}
    2 &SINet~\cite{fan2020camouflaged} &CVPR$_{20}$ 
    &search and identification modules &BTF &- &$\bigstar$ &$\bullet$ &\href{https://github.com/DengPingFan/SINet}{Link}\\
    3 &MirrorNet~\cite{yan2021mirrornet} &Access$_{21}$ 
    & fuse input and mirror data streams &MSF &- &$\bigstar$ &$\bullet$ &\href{https://sites.google.com/view/ltnghia/research/camo}{Link}\\ \rowcolor{mygray2}
    4 &DCE~\cite{xiang2021exploring} &arXiv$_{21}$ 
    & depth contribution exploration, confidence-aware loss &BF &\checkmark &$\bigstar$ &$\bullet$ &\href{https://github.com/JingZhang617/RGBD-COD}{Link} \\
    5 &D2CNet~\cite{wang2022d2cnet} &TIE$_{21}$ 
    &dual-branch, dual-guidance \& cross-refine &BTF &- &$\bigstar$ &$\bullet$ &\href{https://github.com/MS-KangWang/COD-D2Net}{Link} \\ \rowcolor{mygray2}
    6 &C2FNet~\cite{sun2021c2fnet} &IJCAI$_{21}$ 
    &context-aware cross-level fusion &BTF &- &$\bigstar$ &$\bullet$ &\href{https://github.com/thograce/C2FNet}{Link} \\ 
    7 &UR-COD~\cite{kajiura2021improving} &MMAsia$_{21}$ 
    &uncertainty of pseudo-edge labels &MSF &- &$\bigstar$ &$\bullet$ &\href{https://github.com/nobukatsu-kajiura/UR-COD}{Link} \\ \rowcolor{mygray2}
    8 &TINet~\cite{zhu2021inferring} &AAAI$_{21}$ 
    &texture perception \& feature interaction guidance &BF &\checkmark &$\bigstar$ &$\bullet$ &N/A \\ 
    9 &JSCOD~\cite{li2021uncertainty} &CVPR$_{21}$ 
    & uncertainty-aware adversarial learning &MSF &- &$\bigstar$ &$\bullet$ &\href{https://github.com/JingZhang617/Joint_COD_SOD}{Link} \\ \rowcolor{mygray2}
    10 &LSR~\cite{lv2021simultaneously} &CVPR$_{21}$ 
    & localize, segment, \& rank objects simultaneously &BF &\checkmark &$\bigstar$ &$\blacksquare$ &\href{https://github.com/JingZhang617/COD-Rank-Localize-and-Segment}{Link} \\ 
    11 &MGL~\cite{zhai2021mutual} &CVPR$_{21}$ 
    &mutual graph learning &BF &\checkmark &$\bigstar$ &$\bullet$ &\href{https://github.com/fanyang587/MGL}{Link} \\ \rowcolor{mygray2} 
    12 &PFNet~\cite{mei2021camouflaged} &CVPR$_{21}$ 
    & distraction mining, positioning and focus modules &BTF &- &$\bigstar$ &$\bullet$ &\href{https://mhaiyang.github.io/CVPR2021_PFNet/index}{Link} \\  
    13 &UGTR~\cite{yang2021uncertainty} &ICCV$_{21}$ 
    &uncertainty-guided transformer reasoning &BF &\checkmark &$\bigstar$ &$\bullet$ &\href{https://github.com/fanyang587/UGTR}{Link} \\ \rowcolor{mygray2}
    14 &BAS~\cite{qin2022boundary} &arXiv$_{22}$ 
    & residual refinement module, hybrid loss &BTF &- &$\bigstar$ &$\bullet$ &\href{https://github.com/xuebinqin/BASNet}{Link} \\ 
    15 &OSFormer~\cite{pei2022osformer} &ECCV$_{22}$ 
    &location-sensing transformer, coarse-to-fine fusion &BF &\checkmark &$\bigstar$ &$\circ$ &\href{https://github.com/PJLallen/OSFormer}{Link} \\ \rowcolor{mygray2}
    16 &CFL~\cite{le2022camouflaged} &TIP$_{22}$ 
    & camouflage fusion learning &BF &\checkmark &$\bigstar$ &$\circ$ & \href{https://sites.google.com/view/ltnghia/research/camo_plus_plus}{Link} \\ 
    17 &NCHIT~\cite{zhang2022camouflaged} &CVIU$_{22}$ 
    &neighbor connection, hierarchical information transfer &BTF &- &$\bigstar$ &$\bullet$ &N/A \\ \rowcolor{mygray2}
    18 &DTC-Net~\cite{zhai2022deep} &TMM$_{22}$ 
    &local bilinear \& spatial coherence organization &BTF &- &$\bigstar$ &$\bullet$ &N/A \\ 
    19 &C2FNet-V2~\cite{chen2022camouflaged} &TCSVT$_{22}$ 
    & context-aware cross-level fusion &BTF &- &$\bigstar$ &$\bullet$ &\href{https://github.com/Ben57882/C2FNet-TSCVT}{Link} \\ \rowcolor{mygray2}
    20 &CubeNet~\cite{zhuge2022cubenet} &PR$_{22}$ 
    & encoder-decoder framework with X-connection &BF &\checkmark &$\bigstar$ &$\bullet$ &\href{https://github.com/mczhuge/CubeNet}{Link}\\ 
    21 &ERRNet~\cite{ji2022fast} &PR$_{22}$ 
    &selective edge aggregation, reversible re-calibration &BF &\checkmark &$\bigstar$ &$\bullet$ &\href{https://github.com/GewelsJI/ERRNet}{Link} \\ \rowcolor{mygray2}
    22 &TPRNet~\cite{zhang2022tprnet} &TVCJ$_{22}$ 
    &transformer-induced progressive refinement &BTF &- &$\bigstar$ &$\bullet$ &\href{https://github.com/zhangqiao970914/TPRNet}{Link}\\ 
    23 &ANSA-Net~\cite{cheng2022attention} &IJCNN$_{22}$ 
    &attention-based neighbor selective aggregation &BF &\checkmark &$\bigstar$ &$\bullet$ &N/A \\ \rowcolor{mygray2}
    24 &BSANet~\cite{zhu2022can} &AAAI$_{22}$ 
    & boundary-guided separated attention &BF &\checkmark &$\bigstar$ &$\bullet$ &\href{https://github.com/zhuhongwei1999/BSA-Net}{Link} \\ 
    25 &FAPNet~\cite{zhou2022feature} &TIP$_{22}$ 
    & boundary guidance, feature aggregation \& propagation &BF &\checkmark &$\bigstar$ &$\bullet$ &\href{https://github.com/taozh2017/FAPNet}{Link} \\ \rowcolor{mygray2}
    26 &FindNet~\cite{li2022findnet} &TIP$_{22}$ 
    &boundary-and-texture cues (extension of~\cite{zhu2022can}) &BF &\checkmark &$\bigstar$ &$\bullet$ &N/A \\ 
    27 &PINet~\cite{chou2022finding} &ICME$_{22}$ 
    &cascaded decamouflage module, label reweighting &BTF &- &$\bigstar$ &$\bullet$ &N/A \\ \rowcolor{mygray2}
    28 &OCENet~\cite{liu2022modeling} &WACV$_{22}$ 
    &online confidence estimation, dynamic uncertainty loss &BF &\checkmark &$\bigstar$ &$\bullet$ &\href{https://github.com/Carlisle-Liu/OCENet}{Link} \\ 
    29 &BGNet~\cite{sun2022boundary} &IJCAI$_{22}$ 
    &edge-guidance feature \& context aggregation modules &BF &\checkmark &$\bigstar$ &$\bullet$ &\href{https://github.com/thograce/BGNet}{Link} \\ \rowcolor{mygray2}
    30 &PreyNet~\cite{zhang2022preynet} &MM$_{22}$ 
    & bidirectional bridging interaction, predator learning &BF &\checkmark &$\bigstar$ &$\bullet$ &\href{https://github.com/sxu1997/PreyNet}{Link} \\ 
    31 &DTINet~\cite{liu2022boosting} &ICPR$_{22}$ 
    &dual-task interactive transformer &BF &\checkmark &$\bigstar$ &$\bullet$ &\href{https://github.com/liuzywen/COD}{Link} \\ \rowcolor{mygray2}
    32 &ZoomNet~\cite{pang2022zoom} &CVPR$_{22}$ 
    &scale integration \& hierarchical mixed-scale units &MSF &- &$\bigstar$ &$\bullet$ &\href{https://github.com/lartpang/ZoomNet}{Link} \\ 
    33 &FDNet~\cite{zhong2022detecting} &CVPR$_{22}$ 
    & frequency enhancement \& high-order relation modules &MSF &- &$\bigstar$ &$\bullet$ &N/A \\ \rowcolor{mygray2}
    34 &SegMaR~\cite{jia2022segment} &CVPR$_{22}$ 
    & segment, magnify, reiterate in a iterative manner &BTF &- &$\bigstar$ &$\bullet$ &\href{https://github.com/dlut-dimt/SegMaR}{Link} \\ 
    35 &SINetV2~\cite{fan2022concealed} &TPAMI$_{22}$ 
    &neighbor connection decoder, group-reversal attention &BTF &- &$\bigstar$ &$\bullet$ &\href{https://github.com/GewelsJI/SINet-V2}{Link} \\ \rowcolor{mygray2}
    36 &MGL-V2~\cite{zhai2023mglv2} &TIP$_{23}$ 
    &multi-source attention recovery (extension of~\cite{zhai2021mutual}) &BF &\checkmark &$\bigstar$ &$\bullet$ &\href{https://github.com/fanyang587/MGL}{Link} \\ 
    37 &FBNet~\cite{lin2023frequency} &TMCCA$_{23}$ 
    & frequency-aware context aggregation \& attention &BTF &- &$\bigstar$ &$\bullet$ &N/A \\ \rowcolor{mygray2}
    38 &TANet~\cite{ren2023tanet} &TCSVT$_{23}$ 
    &texture-aware refinement, boundary-consistency loss &BTF &- &$\bigstar$ &$\bullet$ &N/A \\ 
    39 &LSR+~\cite{lv2023towards} &TCSVT$_{23}$ 
    & triple task learning (extension of~\cite{lv2021simultaneously}) &BF &\checkmark &$\bigstar$ &$\blacksquare$ &\href{https://github.com/JingZhang617/COD-Rank-Localize-and-Segment}{Link} \\ \rowcolor{mygray2}
    40 &SARNet~\cite{xing2023go} &TCSVT$_{23}$ 
    &triple-stage refinement (search-amplify-recognize) &BTF &- &$\bigstar$ &$\bullet$ &\href{https://github.com/Haozhe-Xing/SARNet}{Link} \\
    41 &MFFN~\cite{zheng2023mffn} &WACV$_{23}$ 
    & co-attention of multi-view, channel fusion unit &MSF &- &$\bigstar$ &$\bullet$ &\href{https://github.com/dwardzheng/MFFN_COD}{Link} \\ \rowcolor{mygray2}
    42 &CRNet~\cite{he2023weakly} &AAAI$_{23}$ 
    &feature-guided and consistency losses &MSF &- &$\diamondsuit$ &$\bullet$ &\href{https://github.com/dddraxxx/Weakly-Supervised-Camouflaged-Object-Detection-with-Scribble-Annotations}{Link} \\ 
    43 &HitNet~\cite{hu2023high} &AAAI$_{23}$ 
    & high-resolution iterative feedback &BTF &- &$\bigstar$ &$\bullet$ &\href{https://github.com/HUuxiaobin/HitNet}{Link} \\ \rowcolor{mygray2} 
    44 &DGNet~\cite{ji2023gradient} &MIR$_{23}$ 
    & gradient-based texture learning, efficient network &BF &\checkmark &$\bigstar$ &$\bullet$ &\href{https://github.com/GewelsJI/DGNet}{Link} \\ 
    45 &FSPNet~\cite{huang2023feature} &CVPR$_{23}$ & feature shrinkage pyramid with transformer &BTF &- &$\bigstar$ &$\bullet$ &\href{https://github.com/ZhouHuang23/FSPNet}{Link} \\
    \rowcolor{mygray2} 
    \rev{46} & \rev{FEDER~\cite{he2023camouflaged}} &\rev{CVPR$_{23}$} & \rev{deep wavelet-like decomposition} &\rev{BTF} &- &\rev{$\bigstar$} &\rev{$\bullet$} &\href{https://github.com/ChunmingHe/FEDER}{Link}\\
    \rev{47} &\rev{DCNet~\cite{luo2023camouflaged}} &\rev{CVPR$_{23}$} &\rev{pixel-level decoupling, instance-level suppression} &\rev{BF} &\rev{\checkmark} &\rev{$\bigstar$} &\rev{$\circ$} &\href{https://github.com/USTCL/DCNet}{Link} \\
    48 &IOCFormer~\cite{sun2023ioc} &CVPR$_{23}$ 
    & unify density- and regression-based strategies &BF &\checkmark &$\bigstar$ &$\square$ &\href{https://github.com/GuoleiSun/Indiscernible-Object-Counting}{Link} \\ 
    49 &PFNet+~\cite{mei2023distraction} &SCIS$_{23}$ 
    &extension of PFNet~\cite{mei2021camouflaged} &BTF &- &$\bigstar$ &$\bullet$ &\href{https://github.com/Mhaiyang/PFNet_Plus}{Link} \\ 
    \rowcolor{mygray2} 
    50 &DQnet~\cite{sun2022dqnet} &arXiv$_{23}$ &cross-model detail querying, relation-based querying &MSF &- &$\bigstar$ &$\bullet$ &\href{https://github.com/CVPR23/DQnet}{Link} \\
    51 &CamoFormer\cite{yin2023camoformer} &arXiv$_{23}$ 
    & masked separable attention &BTF &- &$\bigstar$ &$\bullet$ &\href{https://github.com/HVision-NKU/CamoFormer}{Link} \\ 
    \rowcolor{mygray2} 
    52 &PopNet~\cite{wu2023source} &arXiv$_{23}$ 
    & source-free depth, object pop-out prior &MSF &- &$\bigstar$ &$\bullet$ &\href{https://github.com/Zongwei97/PopNet}{Link} \\
    \hline
\end{tabular}
\end{threeparttable}
\end{table*}

\begin{figure*}
  \centering
  \includegraphics[width=0.8\linewidth]{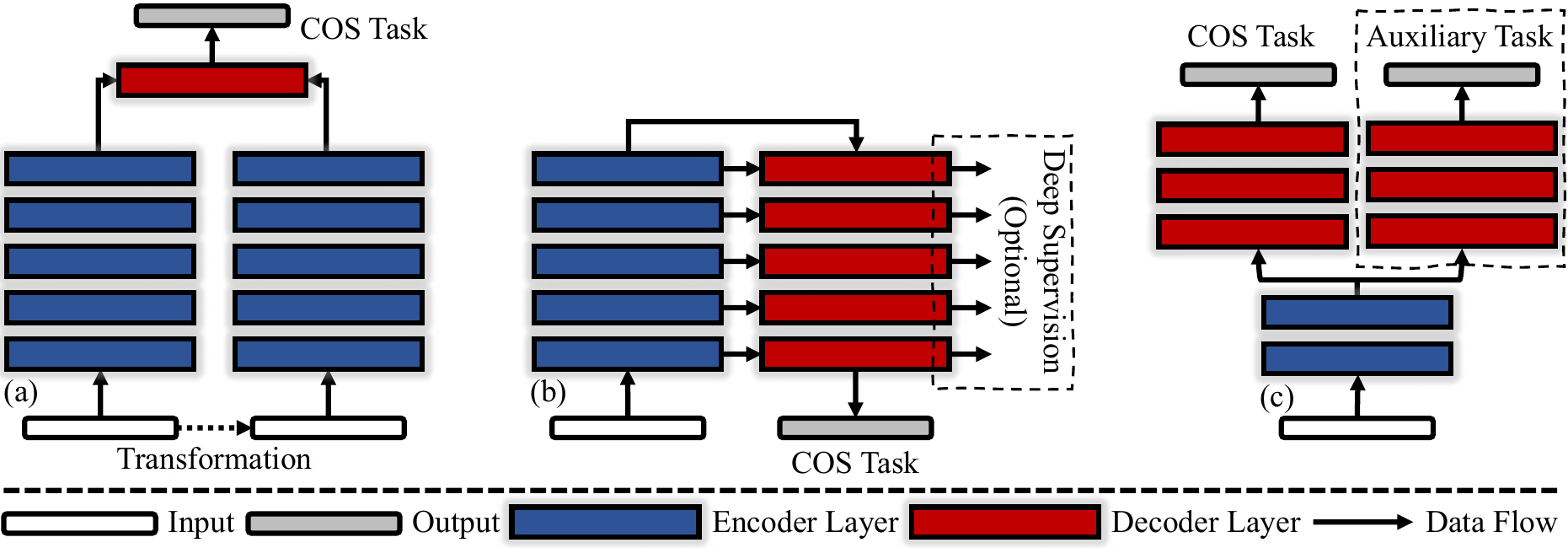}
  \caption{\textbf{Network architectures for COS at a glance.} We present four types of frameworks from left to right: (a) multi-stream framework, (b) bottom-up/top-down framework and its variant with deep supervision (optional), and (c) branched framework. See~\secref{sec:cos_review} for more details.}
  \label{fig:cos_arch_comparison}
\end{figure*}

\subsubsection{Concealed Object Segmentation}\label{sec:cos_review}

This section discusses previous solutions for camouflage object segmentation (COS) from two perspectives: network architecture and {learning paradigm}.

\myPara{Network Architecture.} 
Generally, fully convolutional networks (FCNs~\cite{long2015fully}) are the standard solution for image segmentation as they can receive input of a flexible size and undergo a single feed-forward propagation. As expected, FCN-shaped frameworks dominate the primary solutions for COS, which fall into three categories:
\textbf{\textit{a) Multi-stream framework}}, shown in~\figref{fig:cos_arch_comparison}~(a), contains multiple input streams to learn multi-source representations explicitly. MirrorNet~\cite{yan2021mirrornet} was the first attempt to add an extra data stream as a bio-inspired attack, which can break the camouflaged state. Several recent works have adopted a multi-stream approach to improve their results, such as supplying pseudo-depth generation~\cite{wu2023source}, pseudo-edge uncertainty~\cite{kajiura2021improving}, adversarial learning paradigm~\cite{li2021uncertainty}, frequency enhancement stream~\cite{zhong2022detecting}, multi-scale~\cite{pang2022zoom} or multi-view~\cite{zheng2023mffn} inputs, and multiple backbones~\cite{sun2022dqnet}.
\rev{
Unlike other supervised settings, CRNet~\cite{he2023weakly} is the only weakly-supervised framework that uses scribble labels as supervision. This approach helps to alleviate overfitting problems on limited annotated data.}
\textit{\textbf{b) Bottom-up and top-down framework}},
as shown in~\figref{fig:cos_arch_comparison}~(b), uses deeper features to enhance shallower ones gradually in a single feed-forward pass. 
For example, C2FNet \cite{sun2021c2fnet} adopts this design to improve concealed features from coarse-to-fine levels. In addition, SegMaR \cite{jia2022segment} employs an iterative refinement network with a sub-network based on this strategy.
Furthermore, other studies~\cite{fan2020camouflaged,ren2023tanet,mei2021camouflaged,wang2022d2cnet,fan2022concealed,qin2022boundary,lin2023frequency,zhai2022deep,chen2022camouflaged,yin2023camoformer,hu2023high,zhang2022tprnet,mei2023distraction,cheng2022attention,chou2022finding,zhang2022camouflaged,xing2023go,huang2023feature,he2023camouflaged} utilized a deeply-supervised strategy~\cite{xie2015holistically,lee2015deeply} on various intermediate feature hierarchies using this framework. This practical, also utilized by the feature pyramid network~\cite{lin2017feature}, combines more comprehensive multi-context features through dense top-down and bottom-up propagation and introduces additional supervision signals before final prediction to provide more dependable guidance for deeper layers.
\textit{\textbf{c) Branched framework}}, shown in~\figref{fig:cos_arch_comparison}~(c), is a single-input-multiple-output architecture, consisting of both segmentation and auxiliary task branches. It should be noted that the segmentation part of this branched framework may have some overlap with previous frameworks, such as single-stream~\cite{le2019anabranch} and bottom-up \& top-down~\cite{liu2022modeling,zhang2022preynet,li2021uncertainty,yang2021uncertainty,lv2021simultaneously,lv2023towards,xiang2021exploring,zhai2021mutual,zhuge2022cubenet,ji2022fast,zhou2022feature,zhu2022can,sun2022boundary,ji2023gradient,zhu2021inferring,liu2022boosting,cheng2022attention,zhai2023mglv2,li2022findnet} frameworks. For instance, ERRNet~\cite{ji2022fast} and FAPNet~\cite{zhou2022feature} are typical examples of jointly learning concealed objects and their boundaries. Since these branched frameworks are closely related to the multi-task learning paradigm, we will provide further details.

\myPara{Learning Paradigm.} 
\rev{We discuss two common types of learning paradigms for COS tasks: single-task and multi-task.}
\textit{\textbf{a) Single-task learning}} is the most commonly used paradigm in COS, which involves only a segmentation task for concealed targets. Based on this paradigm, most current works~\cite{fan2020camouflaged,fan2022concealed,chen2022camouflaged} focus on developing attention modules to identify target regions. 
\textit{\textbf{b) Multi-task learning}} introduces an auxiliary task to coordinate or complement the segmentation task, leading to robust COS learning. These multi-task frameworks can be implemented by conducting confidence estimation~\cite{li2021uncertainty,liu2022modeling,zhang2022preynet,yang2021uncertainty}, localization/ranking~\cite{lv2021simultaneously,lv2023towards}, category prediction~\cite{le2019anabranch} tasks and learning depth~\cite{wu2023source,xiang2021exploring}, boundary~\cite{zhai2021mutual,zhuge2022cubenet,ji2022fast,zhou2022feature,zhu2022can,sun2022boundary}, and texture~\cite{ji2023gradient,zhu2021inferring} cues of camouflaged object.

\begin{table*}[ht!]
\centering
\caption{
\textbf{Essential characteristics of reviewed video-level methods.}
\textbf{Optical flow (O.F.):} whether pre-generating optical flow map. \textbf{Supervision level (S.L.):} fully-supervision with real data ($\bigstar$) or synthetic data ($\clubsuit$), and self-supervision ($\heartsuit$). \textbf{Task level (T.L.):} video camouflaged object detection ($\vartriangle$) and segmentation ($\blacktriangle$). 
For further details, refer to \secref{sec:video_level_csu_models}.
}    
\label{tab:video_csu_work_review}
\scriptsize
\renewcommand{\arraystretch}{1}
\renewcommand{\tabcolsep}{0.493cm}
\begin{threeparttable}
\begin{tabular}{| c | r || r | c | ccc | c | }
    \hline
    \rowcolor{mygray1}
    \textbf{\#} &\textbf{Model}~~ &\textbf{Pub.} 
    &\textbf{Core Components} &\textbf{O.F.} &\textbf{S.L.} &\textbf{T.L.}  &\textbf{Project} \\
    \hline
    \hline 
    1 &FMC~\cite{xie2019object} &CVPR$_{19}$ 
    & pixel trajectory recurrent neural network and clustering  & \checkmark &$\bigstar$ & $\blacktriangle$ &N/A \\ \rowcolor{mygray2}
    2 & VRS~\cite{lamdouar2020betrayed} & ACCV$_{20}$ 
    & video registration and motion segmentation network & \checkmark &$\bigstar$ & $\vartriangle$  &\href{https://github.com/hlamdouar/MoCA/}{Link} \\
    3 & SIMO~\cite{Lamdouar21} & BMVC$_{21}$ 
    & dual-head architecture, synthetic dataset  & \checkmark & $\clubsuit$ & $\vartriangle$ &\href{https://www.robots.ox.ac.uk/~vgg/research/simo/}{Link} \\  \rowcolor{mygray2}
    4 & MG~\cite{yang2021selfsupervised} &ICCV$_{21}$ 
    & self-supervised motion grouping  & \checkmark &$\heartsuit$ & $\vartriangle$ &\href{https://github.com/charigyang/motiongrouping}{Link} \\ 
    5 & RCF~\cite{bideau2022right} & arXiv$_{22}$  
    & rotation-compensated flow, camera motion estimation & \checkmark & $\bigstar$ & $\vartriangle$ &N/A \\ \rowcolor{mygray2}
    6 & OCLR~\cite{xie2022segmenting} & NeurIPS$_{22}$ 
    & object-centric layered representation, synthetic dataset & \checkmark & $\clubsuit$ & $\vartriangle$ &N/A \\
    7 & OFS~\cite{meunier2022driven} & TPAMI$_{22}$  
    & expectation-maximization method, motion augmentation & \checkmark & $\heartsuit$ & $\vartriangle$ &\href{https://github.com/Etienne-Meunier-Inria/EM-Flow-Segmentation}{Link} \\ \rowcolor{mygray2}
    8 & QSDI~\cite{kowal2022deeper} &CVPR$_{22}$ 
    & quantifying the static and dynamic biases  & \checkmark & $\bigstar$ & $\vartriangle$ &\href{https://yorkucvil.github.io/Static-Dynamic-Interpretability/}{Link} \\
    9 &SLTNet~\cite{cheng2022implicit} &CVPR$_{22}$ 
    & implicit motion handling, short- and long-term modules & - &$\bigstar$ & $\blacktriangle$ & \href{https://github.com/XuelianCheng/SLT-Net}{Link} \\ 
    \hline
\end{tabular}
\end{threeparttable}
\end{table*}

\subsubsection{Concealed Instance Ranking}

There has been limited research conducted on this topic. Lv~\etal~\cite{lv2021simultaneously} observed for the first time existing COS approaches could not quantify the difficulty level of camouflage. Regarding this issue, they used an eye tracker to create a new dataset, called CAM-LDR~\cite{lv2023towards}, that contains instance segmentation masks, fixation labels, and ranking labels. They also proposed two unified frameworks, LSR~\cite{lv2021simultaneously} and its extension LSR+~\cite{lv2023towards}, to simultaneously learn triple tasks, \ie, localizing, segmenting, and ranking camouflaged objects. The insight behind it is that discriminative localization regions could guide the segmentation of the full scope of camouflaged objects, and then, the detectability of different camouflaged objects could be inferred by the ranking task. 

\subsubsection{Concealed Instance Segmentation}

This task advances the COS task from the regional to the instance level, a relatively new field compared with the COS. Then, Le~\etal~\cite{le2022camouflaged} build a new CIS benchmark, CAMO++, via extending on previous CAMO~\cite{le2019anabranch} dataset. They also proposed a camouflage fusion learning strategy to fine-tune existing instance segmentation models (\eg, Mask R-CNN~\cite{he2017mask}) by learning image contexts. Based on instance benchmarks as in COD10K~\cite{fan2020camouflaged} and NC4K~\cite{lv2021simultaneously}, the first one-stage transformer framework, OSFormer~\cite{pei2022osformer}, was proposed for this field by introducing two core designs: location-sensing transformer and coarse-to-fine fusion. \rev{Recently, Luo~\etal~\cite{luo2023camouflaged} proposed to segment camouflaged instances with two designs: a pixel-level camouflage decoupling module and an instance-level camouflage suppression module.}

\subsubsection{Concealed Object Counting}

Sun~\etal~\cite{sun2023ioc} recently introduced a new challenge for the community called indiscernible object counting (IOC), which involves counting objects that are difficult to distinguish from their surroundings. They created IOCfish5K, a large-scale dataset containing high-resolution images of underwater scenes with many indiscernible objects (focus on fish) and dense annotations to address the lack of appropriate datasets for this challenge. They also proposed a baseline model called IOCFormer by integrating density-based and regression-based methods in a unified framework.

Based on the above summaries, the COS task is experiencing a rapid development period, resulting in numerous contemporary publications each year. However, very few proposed solutions are still proposed for the COL, CIR, and CIS tasks. This suggests that these fields remain under-explored and offer significant room for further research. Notably, many previous studies are available as references (such as saliency prediction~\cite{borji2019saliency}, salient object subitizing~\cite{islam2018revisiting}, and salient instance segmentation~\cite{wu2021regularized}), providing a solid foundation for understanding these tasks from a camouflaged perspective.

\subsection{Video-level CSU Models}\label{sec:video_level_csu_models}

There are two branches for the video-level CSU task, including detecting and segmenting camouflaged objects from videos. Refer~\tabref{tab:video_csu_work_review} for details.

\subsubsection{Video Concealed Object Detection}

Most works~\cite{yang2021selfsupervised,xie2022segmenting} formulated this topic as the degradation problem of the segmentation task since the scarcity of pixel-wise annotations. They, as usual, trained on segmentation datasets (\eg, DAVIS~\cite{perazzi2016benchmark}, FBMS~\cite{ochs2013segmentation}) but evaluated the generalizability performance on video camouflaged object detection dataset, MoCA~\cite{lamdouar2020betrayed}. These methods consistently opt to extract offline optical flow as motion guidance for the segmentation task, but diversifying over the learning strategies, such as fully-supervised learning with real~\cite{lamdouar2020betrayed,bideau2022right,kowal2022deeper} or synthetic~\cite{Lamdouar21,xie2022segmenting} data and self-supervised learning~\cite{yang2021selfsupervised,meunier2022driven}.

\subsubsection{Video Concealed Object Segmentation}

Xie~\etal~\cite{xie2019object} proposed the first work on camouflaged object discovery in videos. They used a pixel-trajectory recurrent neural network to cluster foreground motion for segmentation. However, this work is limited to a small-scale dataset, CAD~\cite{bideau2016s}. Recently, based upon localization-level dataset MoCA~\cite{lamdouar2020betrayed} with bounding box labels, Cheng~\etal~\cite{cheng2022implicit} extended this field by creating a large-scale VCOS benchmark MoCA-Mask with pixel-level masks. They also introduced a two-stage baseline SLTNet to implicitly utilize motion information.

From what we have reviewed above, the current approaches for VCOS tasks are still in a nascent state of development. Several concurrent works in well-established video segmentation fields (\eg, self-supervised correspondence learning~\cite{li2022locality,araslanov2021dense,liu2021emergence,lu2020learning,caron2021emerging}, unified framework for different motion-based tasks~\cite{wang2021different,yan2022towards,xu2022unifying}) points the way to further explore. Besides, considering high-level semantic understanding has a research gap that merits being supplied, such as semantic segmentation and instance segmentation in the camouflaged scenes.

\section{CSU Datasets}\label{sec:dataset}
In recent years, various datasets have been collected for both image- and video-level CSU tasks.  In~\tabref{tab:csu_dataset_review}, we summarize the features of the representative datasets.

\subsection{Image-level Datasets}\label{sec:image_level_datasets}
\textbf{$\bullet$~CAMO-COCO}~\cite{le2019anabranch} 
is tailor-made for COS tasks with 2,500 image samples across eight categories, divided into two sub-datasets, \ie, CAMO with camouflaged objects and MS-COCO with non-camouflaged objects. 
Both CAMO and MS-COCO contain 1,250 images with a split of 1,000 for training and 250 for testing.

\myPara{NC4K}~\cite{lv2021simultaneously} is currently the largest testing set for evaluating COS models. NC4K consists of 4,121 camouflaged images sourced from the Internet and can be divided into two primary categories: natural scenes and artificial scenes. In addition to the images, this dataset also provides localization labels that include both object-level segmentation and instance-level masks, making it a valuable resource for researchers working in this field.
In a recent study by Lv~\etal~\cite{lv2021simultaneously}, an eye tracker was utilized to collect fixation information for each image. As a result, a CAM-FR dataset of 2,280 images was created, with 2,000 images used for training and 280 for testing. The dataset was annotated with three types of labels: localization, ranking, and instance labels.



\myPara{CAMO++}~\cite{le2022camouflaged} is a newly released dataset that contains 5,500 samples, all of which have undergone hierarchical pixel-wise annotation. The dataset is divided into two parts: camouflaged samples (1,700 images for training and 1,000 for testing) and non-camouflaged samples (1,800 images for training and 1,000 for testing).

\myPara{COD10K}~\cite{fan2020camouflaged,fan2022concealed} 
is currently the largest-scale dataset, featuring a wide range of camouflaged scenes. The dataset contains 10,000 images from multiple open-access photography websites, covering ten super-classes and 78 sub-classes. Out of these images, 5,066 are camouflaged, 1,934 are non-camouflaged pictures and 3,000 are background images. The camouflaged subset of COD10K is annotated using different labels such as category labels, bounding boxes, object-level masks, and instance-level masks, providing a diverse set of annotations.

\myPara{CAM-LDR}~\cite{lv2023towards} 
comprises of 4,040 training and 2,026 testing samples. These samples were selected from commonly-used hybrid training datasets (\ie, CAMO with 1,000 training samples and COD10K with 3,040 training samples), along with the testing dataset (\ie, COD10K with 2,026 testing samples). CAM-LDR is an extension of NC4K~\cite{lv2021simultaneously} that includes four types of annotations: localization labels, ranking labels, object-level segmentation masks, and instance-level segmentation masks. The ranking labels are categorized into six difficulty levels -- background, easy, medium1, medium2, medium3, and hard.
 
\myPara{S-COD}~\cite{he2023weakly} is the first dataset designed specifically for the COS task under the weakly-supervised setting. The dataset includes 4,040 training samples, with 3,040 samples selected from COD10K and 1,000 from CAMO. These samples were re-labeled using scribble annotations that provide a rough outline of the primary structure based on first impressions, without pixel-wise ground-truth information.


\myPara{IOCfish5K}~\cite{sun2023ioc} is a distinct dataset that focuses on counting instances of fish in camouflaged scenes. This COC dataset comprises 5,637 high-resolution images collected from YouTube, with 659,024 center points annotated. The dataset is divided into three subsets, with 3,137 images allocated for training, 500 for validation, and 2,000 for testing.


\vspace{.15in}\noindent\textbf{Remarks.} 
In summary, three datasets (CAMO, COD10K, and NC4K) are commonly used as benchmarks to evaluate camouflage object segmentation (COS) approaches, with the experimental protocols typically described in~\secref{sec:cos_experimental_protocols}. For the concealed instance segmentation (CIS) task, two datasets (COD10K and NC4K) containing instance-level segmentation masks can be utilized. The CAM-LDR dataset, which provides fixation information and three types of annotations collected from a physical eye tracker device, is suitable for various brain-inspired explorations in computer vision. Additionally, there are two new datasets from CSU: S-COD, designed for weakly-supervised COS, and IOCfish5K, focused on counting objects within camouflaged scenes.

\subsection{Video-level Datasets}\label{sec:video_level_datasets}
\textbf{$\bullet$~CAD}~\cite{bideau2016s}
is a small dataset comprising nine short video clips and 836 frames. The annotation strategy used in this dataset is sparse, with camouflaged objects being annotated every five frames. As a result, there are 191 segmentation masks available in the dataset.

\myPara{MoCA}~\cite{lamdouar2020betrayed} is a comprehensive video database from YouTube that aims to detect moving camouflaged animals. It consists of 141 video clips featuring 67 categories and comprises 37,250 high-resolution frames with corresponding bounding box labels for 7,617 instances.


\begin{table*}[t!]
\centering
\scriptsize
\caption{\textbf{Essential characteristics for CSU datasets.} 
\textbf{Train/Test:} number of samples for training/testing (\eg, images for image dataset or frames for video dataset)
\textbf{Task:} data type of dataset. \textbf{N.Cam.:} whether collecting non-camouflaged samples. \textbf{Cls.:} whether providing classification labels. \textbf{B.Box:} whether providing bounding box labels for the detection task. \textbf{Obj./Ins.:} whether providing object- or instance-level segmentation masks for segmentation tasks. 
 \textbf{Rank:} whether providing ranking labels for instances. \textbf{Scr.:} whether providing weak labels in scribbled form. \textbf{Cou.:} whether providing dense object counting labels. See \secref{sec:image_level_datasets} and \secref{sec:video_level_datasets} for more descriptions.}
\label{tab:csu_dataset_review}
\renewcommand{\arraystretch}{1}
\renewcommand{\tabcolsep}{0.208cm}
\begin{threeparttable}
\begin{tabular}{| c | r || rr | rr | c | ccccccccc | c |}
    \hline
    \rowcolor{mygray1}
    \textbf{\#} &\textbf{Dataset} & \textbf{Year} &\textbf{Pub.} &\textbf{Train} &\textbf{Test} &\textbf{Task} &\textbf{N.Cam.} &\textbf{Cls.} &\textbf{B.Box} &\textbf{Obj.} &\textbf{Ins.} &\textbf{Fix.} &\textbf{Rank} &\textbf{Scr.} &\textbf{Cou.} & \textbf{Website} \\
    \hline
    \hline
    1 &CAD~\cite{bideau2016s} &2016 &ECCV &0 &836  &Video &- &\checkmark &- &\checkmark &- &- &- &- &- &\href{http://vis-www.cs.umass.edu/motionSegmentation/}{Link} \\ \rowcolor{mygray2}
    2 &CAMO-COCO~\cite{le2019anabranch} &2019 &CVIU &2000 &500 &Image &\checkmark &- &- &\checkmark &- &- &- &- &- &\href{https://sites.google.com/view/ltnghia/research/camo}{Link}\\ 
    3 &MoCA~\cite{lamdouar2020betrayed} &2020 &ACCV &0 &37,250 &Video &- &\checkmark &\checkmark &- &- &- &- &- &- &\href{https://www.robots.ox.ac.uk/~vgg/data/MoCA/}{Link} \\ \rowcolor{mygray2}
    4 &NC4K~\cite{lv2021simultaneously} &2021 &CVPR &0 &4,121 &Image &- &- &\checkmark &\checkmark &\checkmark &- &- &- &- &\href{https://github.com/JingZhang617/COD-Rank-Localize-and-Segment}{Link} \\ 
    5 &MoCA-Mask~\cite{cheng2022implicit} &2022 &CVPR &19,313 &3,626 &Video &- &\checkmark &- &\checkmark &- &- &- &- &- &\href{https://xueliancheng.github.io/SLT-Net-project/}{Link} \\ \rowcolor{mygray2}
    6 &CAMO++~\cite{le2022camouflaged} &2022 &TIP &3,500 &2,000 &Image &\checkmark &- &\checkmark &\checkmark &\checkmark &- &- &- &- &\href{https://sites.google.com/view/ltnghia/research/camo_plus_plus}{Link} \\ 
    7 &COD10K~\cite{fan2020camouflaged,fan2022concealed} &2022 &TPAMI &6,000 &4,000 &Image &\checkmark &\checkmark &\checkmark &\checkmark &\checkmark &- &- &- &- &\href{https://dengpingfan.github.io/pages/COD.html}{Link} \\ \rowcolor{mygray2}
    8 &CAM-LDR~\cite{lv2023towards} &2023 &TCSVT &4,040 &2,026 &Image &- &- &- &\checkmark &\checkmark &\checkmark &\checkmark &- &- &\href{https://github.com/JingZhang617/COD-Rank-Localize-and-Segment}{Link} \\
    9 &S-COD~\cite{he2023weakly} &2023 &AAAI &4,040 &0 &Image &- &- &- &- &- &- &- &\checkmark &- &\href{https://github.com/dddraxxx/Weakly-Supervised-Camouflaged-Object-Detection-with-Scribble-Annotations}{Link} \\ \rowcolor{mygray2}
    10 &Camfish5K~\cite{sun2023ioc} &2023 &CVPR &3,637 &2,000 &Image &- &\checkmark &- &- &- &- &- &- &\checkmark &\href{https://github.com/GuoleiSun/Indiscernible-Object-Counting}{Link} \\
    \hline
\end{tabular}
\end{threeparttable}
\end{table*}

\myPara{MoCA-Mask}~\cite{cheng2022implicit}, 
an extension of MoCA dataset \cite{lamdouar2020betrayed}, provides human-annotated segmentation masks every five frames based on MoCA dataset \cite{lamdouar2020betrayed}. MoCA-Mask is divided into two parts: a training set consisting of 71 short clips (19,313 frames with 3,946 segmentation masks) and an evaluation set containing 16 short clips (3,626 frames with 745 segmentation masks). To label those unlabeled frames, pseudo-segmentation labels were synthesized using a bidirectional optical flow-based strategy~\cite{teed2020raft}.

\vspace{.15in}\noindent\textbf{Remarks.} 
The MoCA dataset is currently the largest collection of videos with concealed objects, while it only offers detection labels. As a result, researchers in the community~\cite{yang2021selfsupervised,xie2022segmenting} typically assess the performance of well-trained segmentation models by converting segmentation masks into detection bounding boxes. Recently, there has been a shift towards video segmentation in concealed scenes with the introduction of MoCA-Mask. Despite these advancements, the quantity and quality of data annotations remain insufficient for constructing a reliable video model that can effectively handle complex concealed scenarios.

\section{CSU Benchmarks}\label{sec:csu_benchmarking}

In this investigation, our benchmarking is built on COS tasks since this topic is relatively well-established and offers a variety of competing approaches. The following sections will detail the evaluation metrics (\secref{sec:cos_evaluation_metrics}), benchmarking protocols (\secref{sec:cos_experimental_protocols}), quantitative analyses (\secref{sec:cos_quan_analysis_camo}, \secref{sec:cos_quan_analysis_nc4k}, \secref{sec:cos_quan_analysis_cod10k}), and qualitative comparisons (\secref{sec:cos_qualitative_comparison}).

\subsection{Evaluation Metrics}\label{sec:cos_evaluation_metrics}
As suggested in~\cite{fan2022concealed}, there are five commonly used metrics\footnote{\url{https://github.com/DengPingFan/CSU/tree/main/cos_eval_toolbox}} available for COS evaluation. We compare a prediction mask $\mathbf{P}$ with its corresponding ground-truth mask $\mathbf{G}$ at the same image resolution.

\myPara{MAE} (mean absolute error, $M$) is a conventional pixel-wise measure, which is defined as:
\begin{equation}\label{equ:mae}
  M = \frac{1}{W \times H} \sum_{x}^{W} \sum_{y}^{H} |\mathbf{P}(x,y) - \mathbf{G}(x,y)|,
\end{equation}
where $W$ and $H$ are the width and height of $\mathbf{G}$, and $(x,y)$ are pixel coordinates in $\mathbf{G}$.

\myPara{F-measure} could be defined as:
\begin{equation}\label{equ:max_f}
  F_\beta = \frac{(1+\beta^2) \textrm{Precision} \times \textrm{Recall}} {\beta^{2} \, \textrm{Precision} + \textrm{Recall}},
\end{equation}
where $\beta^2=0.3$ is used to emphasize precision value over recall value, as recommended in~\cite{borji2015salient}. Other two metrics are derived from:
\begin{equation}
    \textrm{Precision}=\frac{|\mathbf{P}(T) \cap \mathbf{G}|}{|\mathbf{P}(T)|},~\textrm{Recall} =\frac{|\mathbf{P}(T)\cap \mathbf{G}|}{|\mathbf{G}|},
\end{equation}
where $\mathbf{P}(T)$ is a binary mask 
obtained by thresholding the non-binary predicted map $\mathbf{P}$ with a threshold value $T \in [0,255]$. The symbol $|\cdot|$ calculates the total area of the mask inside the map. Therefore, it is possible to convert a non-binary prediction mask into a series of binary masks with threshold values ranging from 0 to 255. By iterating over all thresholds, three metrics are obtained with maximum ($F_{\beta}^{mx}$), mean ($F_{\beta}^{mn}$), and adaptive ($F_{\beta}^{ad}$) values of F-measure.
    
\myPara{Enhanced-alignment measure} ($E_{\phi}$)~\cite{fan2021cognitive,Fan2018Enhanced} is a recently proposed binary foreground evaluation metric, which considers the both local and global similarity between two binary maps. Its formulation is defined as:
\begin{equation}\label{equ:e_m}
E_{\phi} = \frac{1}{W \times H} \sum_{x}^{W} \sum_{y}^{H} \phi\left[\mathbf{P}(x, y), \mathbf{G}(x, y)\right],
\end{equation}
where $\phi$ is the enhanced-alignment matrix. Similar to $F_\beta$, this metric also includes three values computed over all the thresholds, \ie, maximum ($E_{\phi}^{mx}$), mean ($E_{\phi}^{mn}$), and adaptive ($E_{\phi}^{ad}$) values.

\begin{table*}[t!]
\centering
\caption{\textbf{Quantitative comparison on CAMO~\cite{le2019anabranch} testing set.} 
We classify the competing approaches based on two aspects: those using convolution-based backbones such as ResNet~\cite{he2016deep}, Res2Net~\cite{gao2019res2net}, EffNet~\cite{tan2019efficientnet}, and ConvNext~\cite{liu2022convnet}; and those using transformer-based backbones such as MiT~\cite{xie2021segformer}, PVTv2~\cite{wang2022pvt}, and Swin~\cite{liu2021swin}. We test two efficiency metrics, model parameters (Para) and multiply-accumulate operations (MACs), in accordance with the preset input resolution in the original paper. Besides, nine evaluation metrics are reported, and the best three scores are highlighted in \tr{red}, \tg{green}, and \tb {blue}, respectively, with $\uparrow$/$\downarrow$ indicating that higher/lower scores are better. If the results are unavailable since the code has not been public, we use a hyphen (-) to denote it. We will follow these notations in subsequent tables unless otherwise specified.}
\label{tab:cos_benchmark_camo}
\scriptsize
\renewcommand{\arraystretch}{1}
\renewcommand{\tabcolsep}{0.2cm}
\begin{threeparttable}
\begin{tabular}{| r | r || r | crr || c|c|c|ccc|ccc | }
    \hline
    \rowcolor{mygray1}
    \textbf{Model}~ &Pub/Year &Backbone &Input &Para. & MACs &$S_{\alpha}\uparrow$ &$F_\beta^w\uparrow$ &$M\downarrow$ &$E_\phi^{ad}\uparrow$ &$E_\phi^{mn}\uparrow$ &$E_\phi^{mx}\uparrow$ &$F_\beta^{ad}\uparrow$ &$F_\beta^{mn}\uparrow$ &$F_\beta^{mx}\uparrow$ \\
    \hline
    \hline
    \multicolumn{15}{|c|}{\tabincell{c}{$\bullet$~\textbf{Convolution-based Backbone}}} \\ \hline
    SINet~\cite{fan2020camouflaged} &CVPR$_{20}$ &ResNet-50 &$352^2$ &48.95M &19.42G
    &0.745 &0.644 &0.092 &0.825 &0.804 &0.829 &0.712 &0.702 &0.708 \\
    D2CNet~\cite{wang2022d2cnet} &TIE$_{21}$ &Res2Net-50 &320$^2$ &- &-
    &0.774 &0.683 &0.087 &0.844 &0.818 &0.838 &0.747 &0.735 &0.743 \\
    C2FNet~\cite{sun2021c2fnet} &IJCAI$_{21}$ &Res2Net-50 &$352^2$ &28.41M &13.12G
    &0.796 &0.719 &0.080 &0.865 &0.854 &0.864 &0.764 &0.762 &0.771 \\
    TINet~\cite{zhu2021inferring} &AAAI$_{21}$ &ResNet-50 &$352^2$ &28.56M &8.58G
    &0.781 &0.678 &0.087 &0.847 &0.836 &0.848 &0.729 &0.728 &0.745 \\
    JSCOD~\cite{li2021uncertainty} &CVPR$_{21}$ &ResNet-50 &$352^2$ &121.63M &25.20G
    &0.800 &0.728 &0.073 &0.872 &0.859 &0.873 &0.779 &0.772 &0.779 \\
    LSR~\cite{lv2021simultaneously} &CVPR$_{21}$ &ResNet-50 &$352^2$ &57.90M &25.21G
    &0.787 &0.696 &0.080 &0.859 &0.838 &0.854 &0.756 &0.744 &0.753 \\ 
    R-MGL~\cite{zhai2021mutual} &CVPR$_{21}$ &ResNet-50 &$473^2$ &67.64M &249.89G
    &0.775 &0.673 &0.088 &0.848 &0.812 &0.842 &0.738 &0.726 &0.740 \\
    S-MGL~\cite{zhai2021mutual} &CVPR$_{21}$ &ResNet-50 &$473^2$ &63.60M &236.60G
    &0.772 &0.664 &0.089 &0.850 &0.807 &0.842 &0.733 &0.721 &0.739 \\
    PFNet~\cite{mei2021camouflaged} &CVPR$_{21}$ &ResNet-50 &$416^2$ &46.50M &26.54G
    &0.782 &0.695 &0.085 &0.855 &0.841 &0.855 &0.751 &0.746 &0.758 \\
    UGTR~\cite{yang2021uncertainty} &ICCV$_{21}$ &ResNet-50 &$473^2$ &48.87M &127.12G
    &0.785 &0.686 &0.086 &0.861 &0.823 &0.854 &0.749 &0.738 &0.754 \\
    BAS~\cite{qin2022boundary} &arXiv$_{21}$ &ResNet-34 &$288^2$ &87.06M &161.19G
    &0.749 &0.646 &0.096 &0.808 &0.796 &0.808 &0.696 &0.692 &0.703 \\
    NCHIT~\cite{zhang2022camouflaged} &CVIU$_{22}$ &ResNet-50 &288$^2$ &- &- 
    &0.784 &0.652 &0.088 &0.841 &0.805 &0.840 &0.723 &0.707 &0.739 \\
    C2FNet-V2~\cite{chen2022camouflaged} &TCSVT$_{22}$ &Res2Net-50 &352$^2$ &44.94M &18.10G
    &0.799 &0.730 &0.077 &0.869 &0.859 &0.869 &0.777 &0.770 &0.779 \\
    CubeNet~\cite{zhuge2022cubenet} &PR$_{22}$ &ResNet-50 &352$^2$ &- &-
    &0.788 &0.682 &0.085 &0.852 &0.838 &0.860 &0.734 &0.732 &0.750 \\
    ERRNet~\cite{ji2022fast} &PR$_{22}$ &ResNet-50 &352$^2$ &69.76M &20.05G
    &0.779 &0.679 &0.085 &0.855 &0.842 &0.858 &0.731 &0.729 &0.742 \\
    TPRNet~\cite{zhang2022tprnet} &TVCJ$_{22}$ &Res2Net-50 &352$^2$ &32.95M &12.98G
    &0.807 &0.725 &0.074 &0.880 &0.861 &0.883 &0.777 &0.772 &0.785 \\
    FAPNet~\cite{zhou2022feature} &TIP$_{22}$ &Res2Net-50 &352$^2$ &29.52M &29.69G
    &0.815 &0.734 &0.076 &0.877 &0.865 &0.880 &0.776 &0.776 &0.792 \\
    BSANet~\cite{zhu2022can} &AAAI$_{22}$ &Res2Net-50 &384$^2$ &32.58M &29.70G
    &0.794 &0.717 &0.079 &0.866 &0.851 &0.867 &0.768 &0.763 &0.770 \\
    OCENet~\cite{liu2022modeling} &WACV$_{22}$ &ResNet-50 &480$^2$ &60.31M &59.70G
    &0.802 &0.723 &0.080 &0.866 &0.852 &0.865 &0.776 &0.766 &0.777 \\
    BGNet~\cite{sun2022boundary} &IJCAI$_{22}$ &Res2Net-50 &416$^2$ &79.85M &58.45G
    &0.812 &0.749 &0.073 &0.876 &0.870 &0.882 &0.786 &0.789 &0.799 \\
    PreyNet~\cite{zhang2022preynet} &MM$_{22}$ &ResNet-50 &448$^2$ &38.53M &58.10G
    &0.790 &0.708 &0.077 &0.856 &0.842 &0.857 &0.763 &0.757 &0.765 \\
    ZoomNet~\cite{pang2022zoom} &CVPR$_{22}$ &ResNet-50 &384$^2$ &32.38M &95.50G
    &0.820 &0.752 &0.066 &0.883 &0.877 &0.892 &0.792 &0.794 &0.805 \\
    FDNet~\cite{zhong2022detecting} &CVPR$_{22}$ &Res2Net-50 &416$^2$ &- &- 
    &\tg{0.841} &\tg{0.775} &\tb{0.063} &\tb{0.901} &\tb{0.895} &0.908 &\tb{0.803} &\tg{0.807} &\tg{0.826} \\ 
    SegMaR~\cite{jia2022segment} &CVPR$_{22}$ &ResNet-50 &352$^2$ &56.21M &33.63G
    &0.815 &0.753 &0.071 &0.881 &0.874 &0.884 &0.795 &0.795 &0.803 \\
    SINetV2~\cite{fan2022concealed} &TPAMI$_{22}$ &Res2Net-50 &$352^2$ &26.98M &12.28G
    &0.820 &0.743 &0.070 &0.884 &0.882 &0.895 &0.779 &0.782 &0.801 \\ 
    CamoFormer-C\cite{yin2023camoformer} &arXiv$_{23}$ &ConvNeXt-B &384$^2$ &96.69M &50.77G
    &\tr{0.859} &\tr{0.812} &\tr{0.050} &\tr{0.919} &\tr{0.913} &\tr{0.920} &\tr{0.842} &\tr{0.842} &\tr{0.855} \\
    CamoFormer-R\cite{yin2023camoformer} &arXiv$_{23}$ &ResNet-50 &384$^2$ &54.25M &78.85G
    &0.816 &0.712 &0.076 &0.863 &0.874 &\tg{0.916} &0.735 &0.745 &0.813 \\
    PopNet~\cite{wu2023source} &arXiv$_{23}$ &Res2Net-50 &512$^2$ &188.05M &154.88G
    &0.808 &0.744 &0.077 &0.871 &0.859 &0.874 &0.790 &0.784 &0.792 \\
    CRNet~\cite{he2023weakly} &AAAI$_{23}$ &ResNet-50 &320$^2$ &32.65M &11.83G
    &0.735 &0.641 &0.092 &0.829 &0.815 &0.830 &0.709 &0.701 &0.707 \\
    PFNet+~\cite{mei2023distraction} &SCIS$_{23}$ &ResNet-50 &480$^2$ &- &-
    &0.791 &0.713 &0.080 &0.862 &0.850 &0.865 &0.764 &0.761 &0.770 \\
    DGNet-S~\cite{ji2023gradient} &MIR$_{23}$ &EffNet-B1 &$352^2$ &7.02M &2.77G
    &0.826 &0.754 &\tb{0.063} &0.896 &0.893 &0.907 &0.786 &0.792 &0.810 \\
    DGNet~\cite{ji2023gradient} &MIR$_{23}$ &EffNet-B4 &$352^2$ &19.22M &1.20G
    &\tb{0.839} &\tb{0.769} &\tg{0.057} &\tg{0.906} &\tg{0.901} &\tb{0.915} &\tg{0.804} &\tb{0.806} &\tb{0.822} \\
    \hline
    \multicolumn{15}{|c|}{\tabincell{c}{$\bullet$~\textbf{Transformer-based Backbone}}} \\ 
    \hline
    DTINet~\cite{liu2022boosting} &ICPR$_{22}$ &MiT-B5 &256$^2$ &266.33M &144.68G
    &\tb{0.856} &0.796 &\tb{0.050} &\tb{0.918} &\tb{0.916} &\tg{0.927} &0.821 &0.823 &\tb{0.843} \\
    CamoFormer-S\cite{yin2023camoformer} &arXiv$_{23}$ &Swin-B  &384$^2$ &97.27M &64.13G
    &\tr{0.876} &\tr{0.832} &\tr{0.043} &\tr{0.935} &\tr{0.930} &\tr{0.938} &\tr{0.856} &\tr{0.856} &\tr{0.871} \\
    CamoFormer-P\cite{yin2023camoformer} &arXiv$_{23}$ &PVTv2-B4 &384$^2$ &71.40M &39.74G
    &\tg{0.872} &\tg{0.831} &\tg{0.046} &\tg{0.931} &\tg{0.929} &\tr{0.938} &\tg{0.853} &\tg{0.854} &\tg{0.868} \\
    HitNet~\cite{hu2023high} &AAAI$_{23}$ &PVTv2-B2 &704$^2$ &25.73M &55.95G
    &0.849 &\tb{0.809} &0.055 &0.910 &0.906 &\tb{0.910} &\tb{0.833} &\tb{0.831} &0.838 \\
    \hline
\end{tabular}
\end{threeparttable}
\end{table*}

\myPara{Structure measure} ($S_\alpha$)~\cite{fan2017structure,cheng2021structure} is used to measure the structural similarity between a non-binary prediction map and a ground-truth mask:
\begin{equation}\label{equ:s_m}         
  S_\alpha = (1-\alpha)  S_o(\mathbf{P}, \mathbf{G}) + \alpha  S_r(\mathbf{P}, \mathbf{G}),
\end{equation}
where $\alpha$ balances the object-aware similarity $S_o$ and region-aware similarity $S_r$. As in the original paper, we use the default setting for $\alpha=0.5$.

\subsection{Experimental Protocols}\label{sec:cos_experimental_protocols}

Suggested by Fan~\etal~\cite{fan2022concealed}, all competing approaches in the benchmarking were trained on a hybrid dataset comprising the training portions of COD10K~\cite{fan2020camouflaged} and CAMO~\cite{le2019anabranch} datasets, totaling 4,040 samples. The models were then evaluated on three popular used benchmarks: COD10K's testing portion with 2,026 samples~\cite{fan2020camouflaged}, CAMO with 250 samples \cite{le2019anabranch}, and NC4K with 4,121 samples \cite{lv2021simultaneously}.

\subsection{Quantitative Analysis on CAMO}\label{sec:cos_quan_analysis_camo}
As reported in~\tabref{tab:cos_benchmark_camo}, we evaluated 36 deep-based approaches on the CAMO testing dataset~\cite{le2019anabranch} using various metrics. These models were classified into two groups based on the backbones they used: 32 convolutional-based and four transformer-based. As for those models using convolutional-based backbones, several interesting findings are observed: 

\begin{table*}[t!]
\centering
\caption{\textbf{Quantitative comparison on NC4K~\cite{lv2021simultaneously} testing dataset.}}
\label{tab:cos_benchmark_nc4k}
\scriptsize
\renewcommand{\arraystretch}{1}
\renewcommand{\tabcolsep}{0.34cm}
\begin{threeparttable}
\begin{tabular}{| r | r | r || c|c|c|ccc|ccc | }
    \hline
    \rowcolor{mygray1}
    \textbf{Model}~ &Pub/Year &Backbone &$S_{\alpha}\uparrow$ &$F_\beta^w\uparrow$ &$M\downarrow$ &$E_\phi^{ad}\uparrow$ &$E_\phi^{mn}\uparrow$ &$E_\phi^{mx}\uparrow$ &$F_\beta^{ad}\uparrow$ &$F_\beta^{mn}\uparrow$ &$F_\beta^{mx}\uparrow$ \\
    \hline
    \hline
    \multicolumn{12}{|c|}{\tabincell{c}{$\bullet$~\textbf{Convolution-based Backbone}}} \\ \hline
    SINet~\cite{fan2020camouflaged} &CVPR$_{20}$ &ResNet-50 
    &0.808 &0.723 &0.058 &0.883 &0.871 &0.883 &0.768 &0.769 &0.775 \\
    C2FNet~\cite{sun2021c2fnet} &IJCAI$_{21}$ &Res2Net-50 
    &0.838 &0.762 &0.049 &0.901 &0.897 &0.904 &0.788 &0.795 &0.810 \\
    TINet~\cite{zhu2021inferring} &AAAI$_{21}$ &ResNet-50 
    &0.829 &0.734 &0.055 &0.882 &0.879 &0.890 &0.766 &0.773 &0.793 \\
    JSCOD~\cite{li2021uncertainty} &CVPR$_{20}$ &ResNet-50 
    &0.842 &0.771 &0.047 &0.906 &0.898 &0.907 &0.803 &0.806 &0.816 \\
    LSR~\cite{lv2021simultaneously} &CVPR$_{21}$ &ResNet-50 
    &0.840 &0.766 &0.048 &0.904 &0.895 &0.907 &0.802 &0.804 &0.815 \\ 
    R-MGL~\cite{zhai2021mutual} &CVPR$_{21}$ &ResNet-50 
    &0.833 &0.740 &0.052 &0.890 &0.867 &0.893 &0.778 &0.782 &0.800 \\
    S-MGL~\cite{zhai2021mutual} &CVPR$_{21}$ &ResNet-50 
    &0.829 &0.731 &0.055 &0.885 &0.863 &0.893 &0.771 &0.777 &0.797 \\
    PFNet~\cite{mei2021camouflaged} &CVPR$_{21}$ &ResNet-50 
    &0.829 &0.745 &0.053 &0.894 &0.887 &0.898 &0.779 &0.784 &0.799 \\
    UGTR~\cite{yang2021uncertainty} &ICCV$_{21}$ &ResNet-50 
    &0.839 &0.747 &0.052 &0.889 &0.874 &0.899 &0.779 &0.787 &0.807 \\
    BAS~\cite{qin2022boundary} &arXiv$_{21}$ &ResNet-34 
    &0.817 &0.732 &0.058 &0.868 &0.859 &0.872 &0.767 &0.772 &0.782 \\
    NCHIT~\cite{zhang2022camouflaged} &CVIU$_{22}$ &ResNet-50
    &0.830 &0.710 &0.058 &0.872 &0.851 &0.894 &0.751 &0.758 &0.792 \\
    C2FNet-V2~\cite{chen2022camouflaged} &TCSVT$_{22}$ &Res2Net-50 
    &0.840 &0.770 &0.048 &0.900 &0.896 &0.904 &0.799 &0.802 &0.814 \\
    ERRNet~\cite{ji2022fast} &PR$_{22}$ &ResNet-50 
    &0.827 &0.737 &0.054 &0.892 &0.887 &0.901 &0.769 &0.778 &0.794 \\
    TPRNet~\cite{zhang2022tprnet} &TVCJ$_{22}$ &Res2Net-50 
    &0.846 &0.768 &0.048 &0.901 &0.898 &0.911 &0.798 &0.805 &0.820 \\
    FAPNet~\cite{zhou2022feature} &TIP$_{22}$ &Res2Net-50 
    &0.851 &0.775 &0.047 &0.903 &0.899 &0.910 &0.804 &0.810 &0.826 \\
    BSANet~\cite{zhu2022can} &AAAI$_{22}$ &Res2Net-50 
    &0.841 &0.771 &0.048 &0.906 &0.897 &0.907 &0.805 &0.808 &0.817 \\
    OCENet~\cite{liu2022modeling} &WACV$_{22}$ &ResNet-50 
    &0.853 &0.785 &0.045 &0.908 &0.902 &0.913 &0.812 &0.818 &0.831 \\
    BGNet~\cite{sun2022boundary} &IJCAI$_{22}$ &Res2Net-50 
    &0.851 &\tb{0.788} &0.044 &0.911 &0.907 &0.916 &0.813 &0.820 &\tb{0.833} \\
    PreyNet~\cite{zhang2022preynet} &MM$_{22}$ &ResNet-50 
    &0.834 &0.763 &0.050 &0.899 &0.887 &0.899 &0.805 &0.803 &0.811 \\
    ZoomNet~\cite{pang2022zoom} &CVPR$_{22}$ &ResNet-50 
    &0.853 &0.784 &\tb{0.043} &0.907 &0.896 &0.912 &0.814 &0.818 &0.828 \\
    FDNet~\cite{zhong2022detecting} &CVPR$_{22}$ &Res2Net-50 
    &0.834 &0.750 &0.052 &0.895 &0.893 &0.905 &0.774 &0.784 &0.804 \\ 
    SegMaR~\cite{jia2022segment} &CVPR$_{22}$ &ResNet-50 
    &0.841 &0.781 &0.046 &0.905 &0.896 &0.907 &\tb{0.821} &\tb{0.821} &0.826 \\
    SINetV2~\cite{fan2022concealed} &TPAMI$_{22}$ &Res2Net-50 
    &0.847 &0.770 &0.048 &0.901 &0.903 &0.914 &0.792 &0.805 &0.823 \\ 
    CamoFormer-C\cite{yin2023camoformer} &arXiv$_{23}$ &ConvNeXt-B 
    &\tr{0.883} &\tr{0.834} &\tr{0.032} &\tr{0.937} &\tr{0.933} &\tr{0.940} &\tr{0.851} &\tr{0.857} &\tr{0.870} \\
    CamoFormer-R\cite{yin2023camoformer} &arXiv$_{23}$ &ResNet-50 
    &0.855 &\tb{0.788} &\tg{0.042} &\tb{0.913} &0.900 &0.914 &0.820 &\tb{0.821} &0.830 \\
    PopNet~\cite{wu2023source} &arXiv$_{23}$ &Res2Net-50 
    &\tg{0.861} &\tg{0.802} &\tg{0.042} &\tg{0.915} &\tb{0.909} &\tb{0.919} &\tg{0.830} &\tg{0.833} &\tg{0.843} \\
    DGNet-S~\cite{ji2023gradient} &MIR$_{23}$ &EfficientNet-B1 
    &0.845 &0.764 &0.047 &0.902 &0.902 &0.913 &0.789 &0.799 &0.819 \\
    DGNet~\cite{ji2023gradient} &MIR$_{23}$ &EfficientNet-B4 
    &\tb{0.857} &0.784 &\tg{0.042} &0.910 &\tg{0.911} &\tg{0.922} &0.803 &0.814 &\tb{0.833} \\
    \hline
    \multicolumn{12}{|c|}{\tabincell{c}{$\bullet$~\textbf{Transformer-based Backbone}}} \\ 
    \hline
    DTINet~\cite{liu2022boosting} &ICPR$_{22}$ &MiT-B5 
    &0.863 &0.792 &0.041 &\tb{0.914} &0.917 &\tb{0.926} &0.809 &0.818 &0.836 \\
    CamoFormer-S\cite{yin2023camoformer} &arXiv$_{23}$ &Swin-B 
    &\tg{0.888} &\tg{0.840} &\tg{0.031} &\tr{0.941} &\tg{0.937} &\tr{0.946} &\tg{0.857} &\tg{0.863} &\tg{0.877} \\
    CamoFormer-P\cite{yin2023camoformer} &arXiv$_{23}$ &PVTv2-B4 
    &\tr{0.892} &\tr{0.847} &\tr{0.030} &\tr{0.941} &\tr{0.939} &\tr{0.946} &\tr{0.863} &\tr{0.868} &\tr{0.880} \\
    HitNet~\cite{hu2023high} &AAAI$_{23}$ &PVTv2-B2 
    &\tb{0.875} &\tb{0.834} &\tb{0.037} &\tg{0.928} &\tb{0.926} &\tg{0.929} &\tb{0.854} &\tb{0.853} &\tb{0.863} \\
    \hline
\end{tabular}
\end{threeparttable}
\end{table*}

\noindent$\bullet$~CamoFormer-C~\cite{yin2023camoformer} achieved the best performance on CAMO with the ConvNeXt~\cite{liu2022convnet} based backbone, even surpassing some metrics produced by transformer-based methods, such as $S_\alpha$ value: 0.859 (CamoFormer-C) \textit{vs.} 0.856 (DTINet~\cite{liu2022boosting}) \textit{vs.} 0.849 (HitNet~\cite{hu2023high}). However, CamoFormer-R~\cite{yin2023camoformer} with ResNet-50 backbone was unable to outperform competitors with the same backbone, such as using multi-scale zooming (ZoomNet~\cite{pang2022zoom}) and iterative refinement (SegMaR~\cite{jia2022segment}) strategies.

\noindent$\bullet$~As for those Res2Net-based models, FDNet~\cite{zhong2022detecting} achieves the top performance on CAMO with high-resolution input of 416$^2$. Besides, SINetV2~\cite{fan2022concealed} and FAPNet~\cite{zhou2022feature} also achieve satisfactory results using the same backbone but with a small input size of 352$^2$.

\noindent$\bullet$~DGNet~\cite{ji2023gradient}, is an efficient model that stands out with its top\#3 performance compared to heavier models like JSCOD~\cite{li2021uncertainty} (121.63M)  and PopNet~\cite{wu2023source} (181.05M), despite having only 19.22M parameters and 1.20G computation costs. Its performance-efficiency balance makes it a promising architecture for further exploration of its potential capabilities.

\noindent$\bullet$~Interestingly, CRNet~\cite{he2023weakly} -- a weakly-supervised model -- competes favorably with early fully-supervised model SINet~\cite{fan2020camouflaged}. 
It suggests that there is room for developing models to bridge the gap toward better data-efficient learning, \eg, self-/semi-supervised learning.

Furthermore, transformer-based methods significantly improve performance due to their superior long-range modeling capabilities. We here test four transformer-based models on the CAMO testing dataset, yielding three noteworthy findings:

\noindent$\bullet$~CamoFormer-S~\cite{yin2023camoformer}, utilizes a Swin transformer design to enhance the hierarchical modeling ability on concealed content, resulting in superior performance across the entire CAMO benchmark. We also observed that the PVT-based variant CamoFormer-P~\cite{yin2023camoformer} achieves comparable results but with fewer parameters, \ie, 71.40M (CamoFormer-P) \textit{vs.} 97.27M (CamoFormer-R).

\noindent$\bullet$~DTINet~\cite{liu2022boosting} is a dual-branch network that utilizes the MiT-B5 semantic segmentation model from SegFormer~\cite{xie2021segformer} as backbone. Despite having 266.33M parameters, it has not delivered impressive performance due to the challenges of balancing such two heavy branches. Nevertheless, this attempt defies our preconceptions and inspires us to investigate the generalizability of semantic segmentation models in concealed scenarios.

\noindent$\bullet$
We also investigate the impact of input resolution on the performance of different models. HitNet~\cite{hu2023high} uses a high-resolution image of 704$^2$, which can improve the detection of small targets, but at the expense of increased computation costs. Similarly, convolutional-based approaches like ZoomNet~\cite{pang2022zoom} achieved impressive results by taking multiple inputs with different resolutions (the largest being 576$^2$) to enhance segmentation performance. However, not all models benefit from this approach. For instance, PopNet~\cite{wu2023source} with a resolution of 480$^2$ fails to outperform SINetV2~\cite{fan2022concealed} with 352$^2$ in all metrics. This observation raises two critical questions: should high-resolution be used in concealed scenarios, and how can we develop an effective strategy for detecting concealed objects of varying sizes? We will propose potential solutions to these questions and present an interesting analysis of the COD10K in~\secref{sec:cos_quan_analysis_cod10k}.

\subsection{Quantitative Analysis on NC4K}\label{sec:cos_quan_analysis_nc4k}
Compared to the CAMO dataset, the NC4K~\cite{lv2021simultaneously} dataset has a larger data scale and sample diversity, indicating subtle changes may have occurred. \tabref{tab:cos_benchmark_nc4k} presents quantitative results on the current largest COS testing dataset with 4,121 samples. The benchmark includes 28 convolutional-based and four transformer-based approaches. Our observations are:

\noindent$\bullet$~CamoFormer-C~\cite{yin2023camoformer} still outperforms all methods on NC4K. In contrast to the awkward situation observed on CAMO as described in~\secref{sec:cos_quan_analysis_camo}, the ResNet-50 based CamoFormer-R~\cite{yin2023camoformer} now performs better than two other competitors (\ie, SegMaR~\cite{jia2022segment} and ZoomNet~\cite{pang2022zoom}) on NC4K. These results confirm the effectiveness of CamoFormer's decoder design in mapping latent features back to the prediction space, particularly for more complicated scenarios. 

\noindent$\bullet$~DGNet~\cite{ji2023gradient} shows less promise on the challenging NC4K dataset, possibly due to its restricted modeling capability with small model parameters. Nevertheless, this drawback provides an opening for modification since the model has a lightweight and simple architecture.

\noindent$\bullet$~While PopNet~\cite{wu2023source} may not perform well on small-scale CMAO datasets, it has demonstrated potential in challenging NC4K dataset. This indicates that using extra network to synthesize depth priors would be helpful for challenging samples. When compared to SINetV2 based on Res2Net-50~\cite{fan2022concealed}, PopNet has a heavier design (188.05M \textit{vs.} 26.98M) and larger input resolution (512$^2$ \textit{vs.} 352$^2$), but only improves the $E_\phi^{mn}$ value by 0.6\%. 

\noindent$\bullet$~Regarding the CamoFormer~\cite{yin2023camoformer} model, there is now a noticeable difference in performance between its two variants. Specifically, the CamoFormer-S variant based on Swin-B lags behind while the CamoFormer-P variant based on PVTv2-B4 performs better.

\begin{table*}[t!]
\centering
\caption{\textbf{Quantitative comparison on COD10K~\cite{fan2020camouflaged} testing set.} 
}
\label{tab:cos_benchmark_cod10k}
\scriptsize
\renewcommand{\arraystretch}{1}
\renewcommand{\tabcolsep}{0.34cm}
\begin{threeparttable}
\begin{tabular}{| r | r | r || c|c|c|ccc|ccc | }
    \hline
    \rowcolor{mygray1}
    \textbf{Model}~ &Pub/Year &Backbone &$S_{\alpha}\uparrow$ &$F_\beta^w\uparrow$ &$M\downarrow$ &$E_\phi^{ad}\uparrow$ &$E_\phi^{mn}\uparrow$ &$E_\phi^{mx}\uparrow$ &$F_\beta^{ad}\uparrow$ &$F_\beta^{mn}\uparrow$ &$F_\beta^{mx}\uparrow$ \\
    \hline
    \hline
    \multicolumn{12}{|c|}{\tabincell{c}{$\bullet$~\textbf{Convolution-based Backbone}}} \\ \hline
    SINet~\cite{fan2020camouflaged} &CVPR$_{20}$ &ResNet-50
    &0.776 &0.631 &0.043 &0.867 &0.864 &0.874 &0.667 &0.679 &0.691 \\
    D2CNet~\cite{wang2022d2cnet} &TIE$_{21}$ &ResNet-50 
    &0.807 &0.680 &0.037 &0.879 &0.876 &0.887 &0.702 &0.720 &0.736 \\
    C2FNet~\cite{sun2021c2fnet} &IJCAI$_{21}$ &Res2Net-50 
    &0.813 &0.686 &0.036 &0.886 &0.890 &0.900 &0.703 &0.723 &0.743 \\
    TINet~\cite{zhu2021inferring} &AAAI$_{21}$ &ResNet-50 
    &0.793 &0.635 &0.042 &0.848 &0.861 &0.878 &0.652 &0.679 &0.712 \\
    JSCOD~\cite{li2021uncertainty} &CVPR$_{20}$ &ResNet-50 
    &0.809 &0.684 &0.035 &0.882 &0.884 &0.891 &0.705 &0.721 &0.738 \\
    LSR~\cite{lv2021simultaneously} &CVPR$_{21}$ &ResNet-50 
    &0.804 &0.673 &0.037 &0.883 &0.880 &0.892 &0.699 &0.715 &0.732 \\ 
    R-MGL~\cite{zhai2021mutual} &CVPR$_{21}$ &ResNet-50 
    &0.814 &0.666 &0.035 &0.865 &0.852 &0.890 &0.681 &0.711 &0.738 \\
    S-MGL~\cite{zhai2021mutual} &CVPR$_{21}$ &ResNet-50 
    &0.811 &0.655 &0.037 &0.851 &0.845 &0.889 &0.667 &0.702 &0.733 \\
    PFNet~\cite{mei2021camouflaged} &CVPR$_{21}$ &ResNet-50 
    &0.800 &0.660 &0.040 &0.868 &0.877 &0.890 &0.676 &0.701 &0.725 \\
    UGTR~\cite{yang2021uncertainty} &ICCV$_{21}$ &ResNet-50 
    &0.818 &0.667 &0.035 &0.850 &0.853 &0.891 &0.671 &0.712 &0.742 \\
    BAS~\cite{qin2022boundary} &arXiv$_{21}$ &ResNet-34 
    &0.802 &0.677 &0.038 &0.869 &0.855 &0.870 &0.707 &0.715 &0.729 \\
    NCHIT~\cite{zhang2022camouflaged} &CVIU$_{22}$ &ResNet-50 
    &0.792 &0.591 &0.046 &0.794 &0.819 &0.879 &0.596 &0.649 &0.698 \\
    C2FNet-V2~\cite{chen2022camouflaged} &TCSVT$_{22}$ &Res2Net-50 
    &0.811 &0.691 &0.036 &0.890 &0.887 &0.896 &0.718 &0.725 &0.742 \\
    CubeNet~\cite{zhuge2022cubenet} &PR$_{22}$ &ResNet-50 
    &0.795 &0.643 &0.041 &0.862 &0.865 &0.883 &0.669 &0.692 &0.715 \\
    ERRNet~\cite{ji2022fast} &PR$_{22}$ &ResNet-50 
    &0.786 &0.630 &0.043 &0.845 &0.867 &0.886 &0.646 &0.675 &0.702 \\
    TPRNet~\cite{zhang2022tprnet} &TVCJ$_{22}$ &Res2Net-50 
    &0.817 &0.683 &0.036 &0.869 &0.887 &0.903 &0.694 &0.724 &0.748 \\
    FAPNet~\cite{zhou2022feature} &TIP$_{22}$ &Res2Net-50 
    &0.822 &0.694 &0.036 &0.875 &0.888 &0.902 &0.707 &0.731 &0.758 \\
    BSANet~\cite{zhu2022can} &AAAI$_{22}$ &Res2Net-50 
    &0.818 &0.699 &0.034 &0.894 &0.891 &0.901 &0.723 &0.738 &0.753 \\
    OCENet~\cite{liu2022modeling} &WACV$_{22}$ &ResNet-50 
    &0.827 &0.707 &0.033 &0.885 &0.894 &0.905 &0.718 &0.741 &0.764 \\
    BGNet~\cite{sun2022boundary} &IJCAI$_{22}$ &Res2Net-50 
    &0.831 &0.722 &0.033 &0.902 &0.901 &0.911 &0.739 &0.753 &0.774 \\
    PreyNet~\cite{zhang2022preynet} &MM$_{22}$ &ResNet-50 
    &0.813 &0.697 &0.034 &0.894 &0.881 &0.891 &0.731 &0.736 &0.747 \\
    ZoomNet~\cite{pang2022zoom} &CVPR$_{22}$ &ResNet-50 
    &0.838 &\tb{0.729} &\tb{0.029} &0.893 &0.888 &0.911 &\tb{0.741} &\tb{0.766} &0.780 \\
    FDNet~\cite{zhong2022detecting} &CVPR$_{22}$ &Res2Net-50 
    &\tb{0.840} &\tb{0.729} &0.030 &\tb{0.906} &\tg{0.919} &\tr{0.935} &0.728 &0.757 &\tb{0.788} \\ 
    SegMaR~\cite{jia2022segment} &CVPR$_{22}$ &ResNet-50 
    &0.833 &0.724 &0.034 &0.893 &0.899 &0.906 &0.739 &0.757 &0.774 \\
    SINetV2~\cite{fan2022concealed} &TPAMI$_{22}$ &Res2Net-50 
    &0.815 &0.680 &0.037 &0.864 &0.887 &0.906 &0.682 &0.718 &0.752 \\ 
    CamoFormer-C~\cite{yin2023camoformer} &arXiv$_{23}$ &ConvNeXt-B 
    &\tr{0.860} &\tr{0.770} &\tr{0.024} &\tr{0.926} &\tr{0.926} &\tr{0.935} &\tr{0.778} &\tr{0.798} &\tr{0.818} \\
    CamoFormer-R~\cite{yin2023camoformer} &arXiv$_{23}$ &ResNet-50 
    &0.838 &0.724 &\tb{0.029} &0.900 &\tb{0.916} &\tg{0.930} &0.721 &0.753 &0.786 \\
    PopNet~\cite{wu2023source} &arXiv$_{23}$ &Res2Net-50 
    &\tg{0.851} &\tg{0.757} &\tg{0.028} &\tg{0.910} &0.910 &\tb{0.919} &\tg{0.771} &\tg{0.786} &\tg{0.802} \\
    CRNet~\cite{he2023weakly} &AAAI$_{23}$ &ResNet-50 
    &0.733 &0.576 &0.049 &0.845 &0.832 &0.845 &0.637 &0.633 &0.636 \\
    PFNet+~\cite{mei2023distraction} &Ssis$_{23}$ &ResNet-50 
    &0.806 &0.677 &0.037 &0.880 &0.884 &0.895 &0.698 &0.716 &0.734 \\
    DGNet-S~\cite{ji2023gradient} &MIR$_{23}$ &EfficientNet-B1 
    &0.810 &0.672 &0.036 &0.869 &0.888 &0.905 &0.680 &0.710 &0.743 \\
    DGNet~\cite{ji2023gradient} &MIR$_{23}$ &EfficientNet-B4 
    &0.822 &0.693 &0.033 &0.879 &0.896 &0.911 &0.698 &0.728 &0.759 \\
    \hline
    \multicolumn{12}{|c|}{\tabincell{c}{$\bullet$~\textbf{Transformer-based Backbone}}} \\ 
    \hline
    DTINet~\cite{liu2022boosting} &ICPR$_{22}$ &MiT-B5 
    &0.824 &0.695 &\tb{0.034} &0.881 &0.896 &0.911 &0.702 &0.726 &0.754 \\
    CamoFormer-S\cite{yin2023camoformer} &arXiv$_{23}$ &Swin-B 
    &\tb{0.862} &\tb{0.772} &\tg{0.024} &\tg{0.932} &\tb{0.931} &\tr{0.941} &\tb{0.780} &\tb{0.799} &\tb{0.818} \\
    CamoFormer-P\cite{yin2023camoformer} &arXiv$_{23}$ &PVTv2-B4 
    &\tg{0.869} &\tg{0.786} &\tr{0.023} &\tb{0.931} &\tg{0.932} &\tg{0.939} &\tg{0.794} &\tg{0.811} &\tg{0.829} \\
    HitNet~\cite{hu2023high} &AAAI$_{23}$ &PVTv2-B2 
    &\tr{0.871} &\tr{0.806} &\tr{0.023} &\tr{0.936} &\tr{0.935} &\tb{0.938} &\tr{0.818} &\tr{0.823} &\tr{0.838} \\
    \hline
\end{tabular}
\end{threeparttable}
\end{table*}

\subsection{Quantitative Analysis on COD10K}\label{sec:cos_quan_analysis_cod10k}
In~\tabref{tab:cos_benchmark_cod10k}, we present a performance comparison of 36 competitors, including 32 convolutional-based models and four transformer-based models, on the COD10K dataset with diverse concealed samples. Based on our evaluation, we have made the following observations:

\noindent$\bullet$~CamoFormer-C~\cite{yin2023camoformer}, which has a robust backbone, remains the best-performing method among all convolutional-based methods. Similarly to its performance on NC4K, CamoFormer-R~\cite{yin2023camoformer} has once again outperformed strong competitors with identical backbones such as SegMaR \cite{jia2022segment} and ZoomNet \cite{pang2022zoom}.

\noindent$\bullet$~
Similar to its performance on the NC4K dataset, PopNet~\cite{wu2023source} achieves consistently high results on the COD10K dataset, ranking second only to CamoFormer-C~\cite{yin2023camoformer}. We believe that prior knowledge of the depth of the scene plays a crucial role in enhancing the understanding of concealed environments. This insight will motivate us to investigate more intelligent ways to learn structural priors, such as incorporating multi-task learning or heuristic methods into our models.

\noindent$\bullet$~Notably, HitNet~\cite{hu2023high} achieves the highest performance on the COD10K benchmark, outperforming models with stronger backbones like Swin-B and PVTv2-B4. To understand why this is the case, we calculated the average resolution of all samples in the CAMO (W=693.89 and H=564.22), NC4K (W=709.19 and H=529.61), and COD10K (W=963.34 and H=740.54) datasets. We found that the testing set for COD10K has the highest overall resolution, which suggests that models utilizing higher resolutions or multi-scale modeling would benefit from this characteristic. Therefore, HitNet is an excellent choice for detecting concealed objects in scenarios where high-resolution images are available.

\subsection{Qualitative Comparison}\label{sec:cos_qualitative_comparison}

\begin{figure*}[t!]
  \centering
  \includegraphics[width=0.99\linewidth]{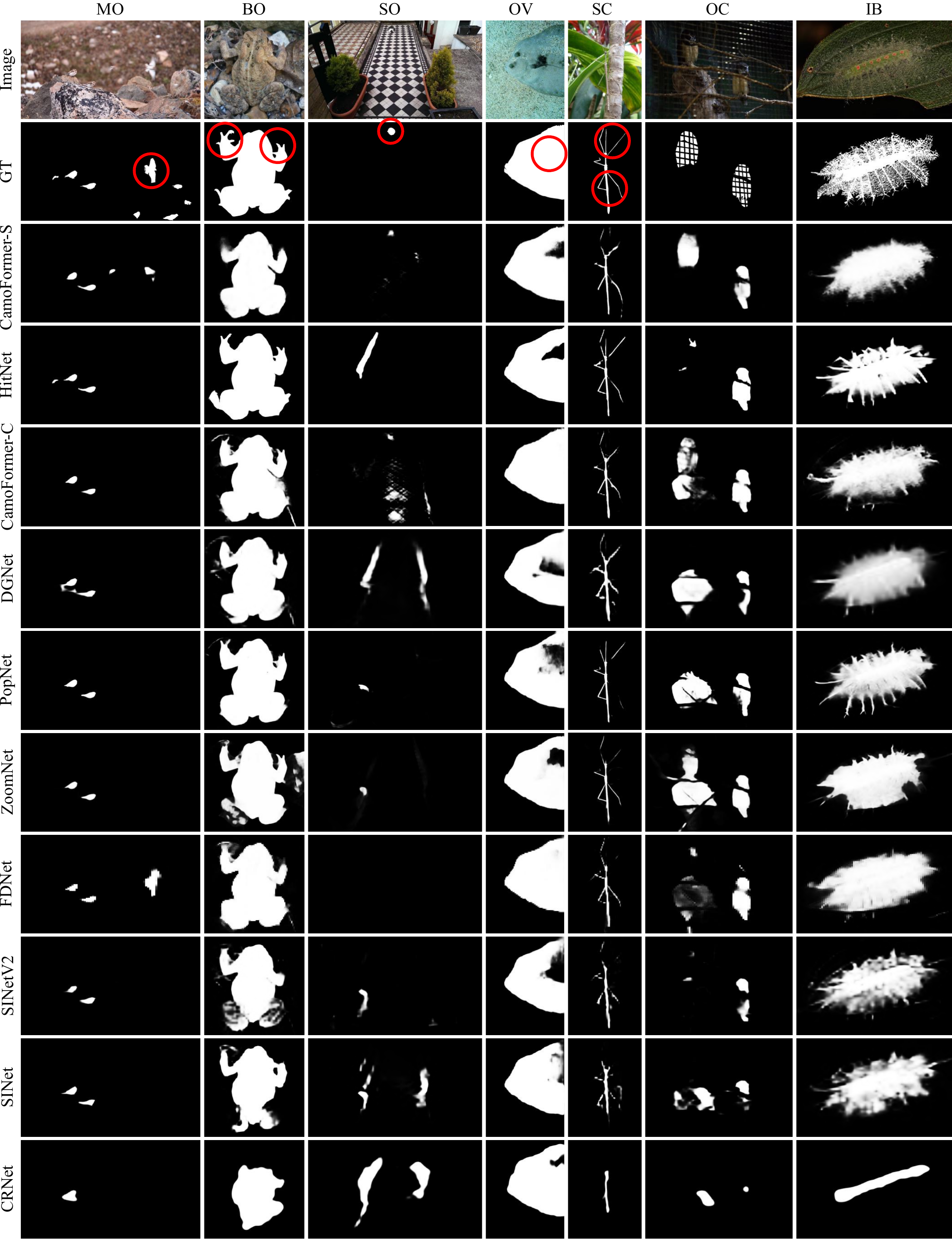}
  \caption{\textbf{Qualitative results of ten COS approaches.} More descriptions on visual attributes in each column refer to~\secref{sec:cos_qualitative_comparison}.}
  \label{fig:cos_quali_viz}
\end{figure*}

This section visually assesses the performance of current top models on challenging and complex samples that are prone to failure. We compare qualitative results predicted by ten groups of top-performing models, including six convolutional-based models (\ie, CamoFormer-C \cite{yin2023camoformer}, DGNet \cite{ji2023gradient}, PopNet \cite{wu2023source}, ZoomNet \cite{pang2022zoom}, FDNet \cite{zhong2022detecting} and SINetV2 \cite{fan2022concealed}), two transformer-based models (\ie, CamoFormer-S \cite{yin2023camoformer} and HitNet\cite{hu2023high}), as well as two other competitors (\ie, the earliest baseline SINet \cite{fan2020camouflaged} and a weakly-supervised model CRNet \cite{he2023weakly}). All samples are selected from the COD10K testing dataset according to seven fine-grained attributes.
The qualitative comparison is presented in \figref{fig:cos_quali_viz}, revealing several interesting findings.

\noindent$\bullet$~The attribute of multiple objects (MO) poses a challenge due to the high false-negative rate in current top-performing models. As depicted in the first column of \figref{fig:cos_quali_viz}, only two out of ten models could locate the white flying bird approximately, as indicated by the red circle in the GT mask. These two models are CamoFormer-S \cite{yin2023camoformer}, which employs a robust transformer-based encoder, and FDNet \cite{zhong2022detecting}, which utilizes a frequency domain learning strategy.

\noindent$\bullet$~The models we tested can accurately detect big objects (BO) by precisely locating the target's main part. However, these models struggle to identify smaller details such as the red circles highlighting the toad's claws in the second column of \figref{fig:cos_quali_viz}.

\noindent$\bullet$~Small object (SO) attribute presents a challenge as it only occupies a small area in the image, typically less than 10\% of the total pixels as reported by COD10K~\cite{fan2020camouflaged}. As shown in the third column of \figref{fig:cos_quali_viz}, only two models (CamoFormer-S and CamoFormer-C \cite{yin2023camoformer}) can detect a cute cat lying on the ground in the distance. Such a difficulty arises for two main reasons: firstly, models struggle to differentiate small objects from complex backgrounds or other irrelevant objects in an image; secondly, detectors may miss small regions due to down-sampling operations caused by low-resolution inputs.

\noindent$\bullet$~Out-of-view (OV) attribute refers to objects partially outside the image boundaries, leading to incomplete representation. To address this issue, a model should have a better holistic understanding of the concealed scene. As shown in the fourth column of \figref{fig:cos_quali_viz}, both CamoFormer-C~\cite{yin2023camoformer} and FDNet~\cite{zhong2022detecting} can handle the OV attribute and maintain the object's integrity. However, two transformer-based models failed to do so. This observation has inspired us to explore more efficient methods, such as local modeling within convolutional frameworks and cross-domain learning strategies.

\noindent$\bullet$~Shape complexity (SC) attribute indicates that an object contains thin parts, such as an animal's foot. In the fifth column of \figref{fig:cos_quali_viz}, the stick insect's feet are a good example of this complexity, being elongated and slender and thus difficult to predict accurately. Only HitNet~\cite{hu2023high} with high-resolution inputs can predict a right-bottom foot (indicated by a red circle).

\noindent$\bullet$~The attribute of occlusion (OC) refers to the partial occlusion of objects, which is a common challenge in general scenes~\cite{qi2022occluded}. In \figref{fig:cos_quali_viz}, for example, the sixth column shows two owls partially occluded by a wire fence, causing their visual regions to be separated. Unfortunately, most of the models presented were unable to handle such cases.

\noindent$\bullet$~Indefinable boundary (IB) attribute is hard to address since its uncertainty between foreground and background. As shown in the last column of \figref{fig:cos_quali_viz}, a matting-level sample.

\noindent$\bullet$~In the last two rows of \figref{fig:cos_quali_viz}, we display the predictions generated by SINet~\cite{fan2020camouflaged}, which was our earliest baseline model. Current models have significantly improved location accuracy, boundary details, and other aspects. Additionally, CRNet \cite{he2023weakly}, a weakly-supervised method with only weak label supervision, can effectively locate target objects to meet satisfactory standards.

\section{Discussion and Outlook}\label{sec:future_research_perspective}
Based on our literature review and experimental analyses, we discuss five challenges and potential CSU-related directions in this section.  

\myPara{Annotation-Efficient Learning.} 
Deep learning techniques have significantly advanced the field of CSU. However, 
conventional supervised deep learning is data-hungry and resource-consuming. In practical scenarios, we hope the models can work on limited resources and have good generalizability. 
Thus developing effective learning strategies 
for CSU tasks is a promising direction, \eg, weakly-supervised strategy in CRNet~\cite{he2023weakly}.

\myPara{Domain Adaptation.}
Camouflaged samples are generally collected from natural scenes. Thus, deploying the models to detect concealed objects in auto-driving scenarios is challenging.
Recent practice demonstrates that various techniques can be used to alleviate this problem, \eg, domain adaptation~\cite{wang2018deep}, transfer learning~\cite{zhuang2020comprehensive}, few-shot learning~\cite{wang2020generalizing}, and meta-learning~\cite{hospedales2021meta}. 

\myPara{High-Fidelity Synthetic Dataset.}
To alleviate algorithmic biases, increasing the diversity and scale of data is crucial. The rapid development of AI-generated content (AIGC)\cite{cao2023comprehensive} and deep generative models, such as generative adversarial networks\cite{goodfellow2020generative,radford2016unsupervised,karras2018progressive} and diffusion models~\cite{rombach2022high,zhang2023adding}, is making it easier to create synthetic data for general domains.
Recently, to address the scarcity of multi-pattern training images, Luo~\etal~\cite{luo2023camdiff} proposed a diffusion-based image generation framework that generates salient objects on a camouflaged sample while preserving its original label. Therefore, a model should be capable of distinguishing between camouflaged and salient objects to achieve a robust feature representation.


\myPara{Neural Architecture Search.} 
Automatic network architecture search (NAS) is a promising research direction that can discover optimal network architectures for superior performance on a given task. In the context of concealment, NAS can identify more effective network architectures to handle complex background scenes, highly variable object appearances, and limited labeled data. This can lead to the developing of more efficient and effective network architectures, resulting in improved accuracy and efficiency.
Combining NAS with other research directions, such as domain adaptation and data-efficient learning, can further enhance the understanding of concealed scenes. These avenues of exploration hold significant potential for advancing the state-of-the-art and warrant further investigation in future research.


\begin{figure*}
  \centering
  \includegraphics[width=.98\linewidth]{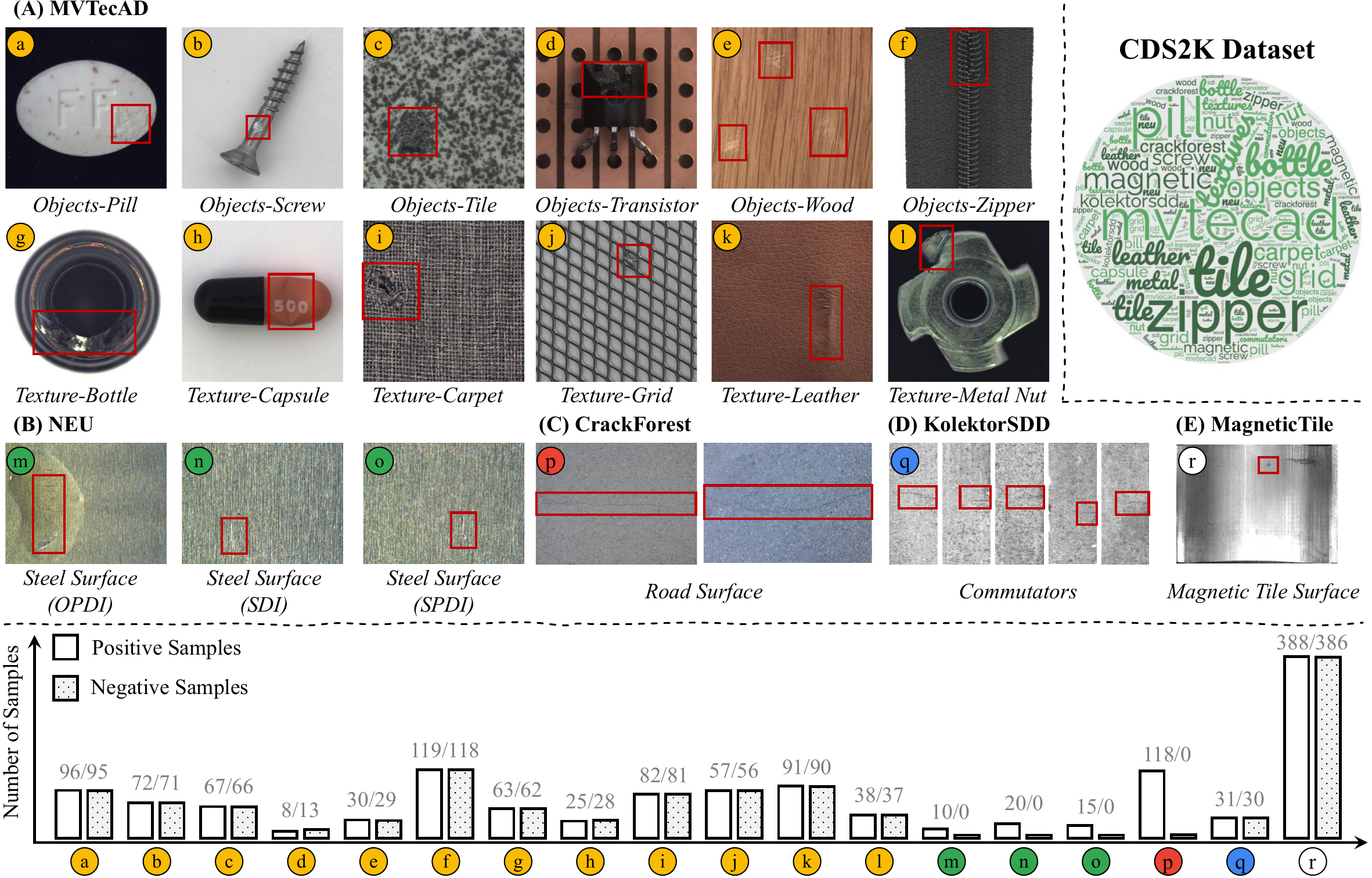}
  \caption{\textbf{Sample gallery of our \ourDataset.} It is collected from five sub-databases: (a-l) MVTecAD, (m-o) NEU, (p) CrackForest, (q) KolektorSDD, and (r) MagneticTile. The defective regions are highlighted with red rectangles. (Top-Right) Word cloud visualization of \ourDataset. (Bottom) The statistic number of positive/negative samples of each category in our \ourDataset.}
  \label{fig:cds2k_samples}
\end{figure*}

\myPara{Large Model and Prompt Engineering.} 
This topic has gained popularity and has even become a direction for the natural language processing community. Recently, the Segment Anything Model (SAM)~\cite{kirillov2023segment} has revolutionized computer vision algorithms, although it has limitations~\cite{ji2023sam} in unprompted settings on several concealed scenarios.
One can leverage the prompt engineering paradigm to simplify workflows using a well-trained robust encoder and task-specific adaptions, such as task-specific prompts and multi-task prediction heads. This approach is expected to become a future trend within the computer vision community.
Large language models (LLMs) have brought both new opportunities and challenges to AI, moving towards artificial general intelligence further. However, it is challenging for academia to train the resource-consuming large models. There could be a promising paradigm that the state-of-the-art deep CSU models are used as the domain experts, and meanwhile, the large models could work as an external component to assist the expert models by providing an auxiliary decision, representation, \etc

\section{Defect Segmentation Dataset}\label{sec:defect_seg_dataset}
Industrial defects usually originate from the undesirable production process, \eg, mechanical impact, workpiece friction, chemical corrosion, and other unavoidable physical, whose external visual form is usually with unexpected patterns or outliers, \eg, surface scratches, spots, holes on industrial devices; color difference, indentation on fabric surface; impurities, breakage, stains on the material surface, \etc~
Though previous works achieve promising advances for identifying visual defects by vision-based techniques, such as classification~\cite{masci2012steel,malhi2004pca,luo2020automated}, detection~\cite{ngan2011automated,kumar2008computer,ghorai2012automatic}, and segmentation~\cite{bergmann2018improving,tabernik2020segmentation,tsai2021auto}.
These techniques work on the assumption that defects are easily detected, but they ignore those challenging defects that are ``seamlessly'' embedded in their materials surroundings. With this, we elaborately \rev{collect} a new multi-scene benchmark, named \ourDataset, for the concealed defect segmentation task, whose samples are selected from existing industrial defect databases.

\subsection{Dataset Organisation}
To create a dataset of superior quality, we established three principles for selecting data: 
(a) The chosen sample should include at least one defective region, which will serve as a positive example.
(b) The defective regions should have a pattern similar to the background, making them difficult to identify.
(c) We also select normal cases as negative examples to provide a contrasting perspective with the positive ones. These samples were selected from the following well-known defect segmentation databases.

\noindent$\bullet$~MVTecAD\footnote{\url{https://www.mvtec.com/company/research/datasets/mvtec-ad}}~\cite{bergmann2021mvtec,bergmann2019mvtec} contains several positive and negative samples for unsupervised anomaly detection. We manually select 748 positive and 746 negative samples with concealed patterns from two main categories: (a) object category as in the 1$^{st}$ row of~\figref{fig:cds2k_samples}: pill, screw, tile, transistor, wood, and zipper. (b) texture category as in the 2$^{nd}$ row of~\figref{fig:cds2k_samples}: bottle, capsule, carpet, grid, leather, and metal nut. The number of positive/negative samples is shown with yellow circles in~\figref{fig:cds2k_samples}
        
\noindent$\bullet$~NEU\footnote{\url{http://faculty.neu.edu.cn/songkechen/zh_CN/zdylm/263270/list/index.htm}} provides three different database: oil pollution defect images~\cite{song2014surface} (OPDI), spot defect images~\cite{bao2021triplet} (SDI), and steel pit defect images~\cite{he2019end} (SPDI). As shown in the third row (green circles) of~\figref{fig:cds2k_samples}, we select 10, 20, and 15 positive samples from these databases separately.
        
\noindent$\bullet$~CrackForest\footnote{\url{https://github.com/cuilimeng/CrackForest-dataset}}~\cite{shi2016automatic,cui2015pavement} is a densely-annotated road crack image database for the health monitoring of urban road surface. We select 118 samples with concealed patterns from them, and the samples are shown in the third row (red circle) of~\figref{fig:cds2k_samples}.

\noindent$\bullet$~KolektorSDD\footnote{\url{https://www.vicos.si/resources/kolektorsdd/}}~\cite{tabernik2020segmentation} collected and annotated by Kolektor Group, which contains several \rev{defective and} non-defective surfaces from the controlled industrial environment in a real-world case. We manually select 31 positive and 30 negative samples with concealed patterns, and the samples are shown in the third row (blue circle) of~\figref{fig:cds2k_samples}.

\noindent$\bullet$~Magnetic Tile Defect\footnote{\url{https://github.com/abin24/Magnetic-tile-defect-datasets}}~\cite{huang2020surface} datasets contains six common magnetic tile defects and corresponding dense annotations. 
We picked 388 positive and 386 negative examples, displayed as white circles in~\figref{fig:cds2k_samples}.

\begin{table*}[t!]
\centering
\scriptsize
\caption{\textbf{Statistic of positive samples in CDS2K.} The region ratio is calculated by $r=\text{defective pixels}/\text{all pixels}$ for a given image. Of note, we only count the number of positive samples in five sub-datasets.}
\label{tab:data_statistic}
\renewcommand{\arraystretch}{1}
\renewcommand{\tabcolsep}{0.346cm}
\begin{threeparttable}
\begin{tabular}{| c | r || ccc ccc | r |}
    \hline
    \rowcolor{mygray1}
    \multicolumn{2}{|c||}{\tabincell{c}{Category}} & $0\% < r < 1\%$ &$1\% \leq r < 10\%$ &$10\% \leq r < 20\%$ &$20\% \leq r < 30\%$ &$30\% \leq r < 40\%$ &$40\% \leq r < 50\%$ & \textbf{Total} \\
    \hline
    \hline
    \multirow{12}{*}{\begin{sideways}MVTecAD\end{sideways}}
    &Objects-Pill &41 &55 &0 &0 &0 &0 &\textbf{96} \\
    &Objects-Screw &71 &1 &0 &0 &0 &0 &\textbf{72} \\
    &Objects-Tile &0 &30 &28 &7 &2 &0 &\textbf{67} \\
    &Objects-Transistor &1 &7 &0 &0 &0 &0 &\textbf{8} \\
    &Objects-Wood &2 &26 &2 &0 &0 &0 &\textbf{30} \\
    &Objects-Zipper &16 &102 &1 &0 &0 &0 &\textbf{119} \\
    &Texture-Bottle &3 &39 &20 &1 &0 &0 &\textbf{63} \\
    &Texture-Capsule &17 &8 &0 &0 &0 &0 &\textbf{25} \\
    &Texture-Carpet &37 &45 &0 &0 &0 &0 &\textbf{82} \\
    &Texture-Grid &39 &18 &0 &0 &0 &0 &\textbf{57} \\
    &Texture-Leather &70 &21 &0 &0 &0 &0 &\textbf{91} \\
    &Texture-Metal Nut &6 &31 &1 &0 &0 &0 &\textbf{38} \\
    \hline
    \multirow{3}{*}{\begin{sideways}NEU\end{sideways}}
    & OPDI  &10 &0 &0 &0 &0 &0 &\textbf{10} \\
    & SDI &20 &0 &0 &0 &0 &0 &\textbf{20} \\
    & SPDI &15 &0 &0 &0 &0 &0 &\textbf{15} \\
    \hline
    \multicolumn{2}{|r||}{\tabincell{l}{CrackForest}} &28 &90 &0 &0 &0 &0 &\textbf{118} \\
    \hline
    \multicolumn{2}{|r||}{\tabincell{l}{KolektorSDD}} &31 &0 &0 &0 &0 &0 &\textbf{31} \\
    \hline
    \multicolumn{2}{|r||}{\tabincell{l}{Magnetic Tile Defect}} &216 &70 &27 &27 &24 &24 &\textbf{388} \\
    \hline
    \multicolumn{2}{|r||}{\tabincell{l}{\textbf{Total}}} &\textbf{623} &\textbf{543} &\textbf{79} &\textbf{35} &\textbf{26} &\textbf{24} &\textbf{1330} \\
    \hline
\end{tabular}
\end{threeparttable}
\end{table*}

\begin{table*}[t!]
\centering
\caption{\textbf{Quantitative comparison on the positive samples of CDS2K.} 
}
\label{tab:cds_zero_performance}
\scriptsize
\renewcommand{\arraystretch}{1}
\renewcommand{\tabcolsep}{0.343cm}
\begin{threeparttable}
\begin{tabular}{| r | r | r || c|c|c|ccc|ccc | }
    \hline
    \rowcolor{mygray1}
    \textbf{Model}~ &Pub/Year &Backbone &$S_{\alpha}\uparrow$ &$F_\beta^w\uparrow$ &$M\downarrow$ &$E_\phi^{ad}\uparrow$ &$E_\phi^{mn}\uparrow$ &$E_\phi^{mx}\uparrow$ &$F_\beta^{ad}\uparrow$ &$F_\beta^{mn}\uparrow$ &$F_\beta^{mx}\uparrow$ \\
    \hline
    \hline
    SINetV2~\cite{fan2022concealed} &TPAMI$_{22}$ &Res2Net-50 &0.551 &0.215 &0.102 &0.509 &0.567 &0.597 &0.223 &0.248 &0.258 \\
    HitNet~\cite{hu2023high} &AAAI$_{23}$ &PVTv2-B2 &0.563 &0.276 &0.118 &0.574 &0.564 &0.570 &0.298 &0.298 &0.299 \\
    DGNet~\cite{ji2023gradient} &MIR$_{23}$ &EfficientNet-B4 &0.578 &0.258 &0.089 &0.552 &0.569 &0.579 &0.274 &0.291 &0.297 \\
    CamoFormer-P\cite{yin2023camoformer} &arXiv$_{23}$ &PVTv2-B4 &0.589 &0.298 &0.100 &0.590 &0.588 &0.596 &0.330 &0.329 &0.339 \\
    \hline
\end{tabular}
\end{threeparttable}
\end{table*}

\begin{figure}
  \centering
  \includegraphics[width=\linewidth]{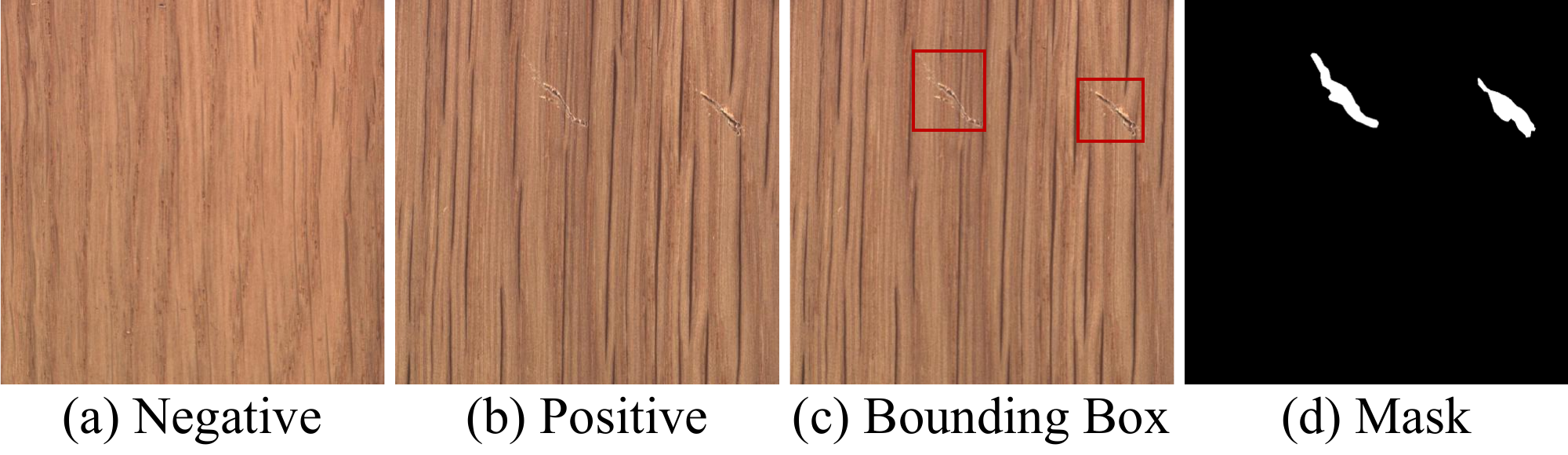}
  \caption{\textbf{Visualization of different annotations.} We select a group of images from the MVTecAD database, including a negative (a) and a positive (b) sample. Corresponding annotations are provided: category (scratches on wood) and defect locations: bounding box (c) and segmentation mask (d).}
  \label{fig:cds2k_annotation}
\end{figure}

\subsection{Dataset Description}

The \ourDataset~comprises 2,492 samples, consisting of 1,330 positive and 1,162 negative instances. Three different human-annotated labels are provided to each sample -- category, bounding box, and pixel-wise segmentation mask. \figref{fig:cds2k_annotation} illustrates examples of these annotations. The average ratio of defective regions for each category is presented in \tabref{tab:data_statistic}, which indicates that most of the defective regions are relatively small.

\subsection{Evaluation on \ourDataset}
Here, we evaluate the generalizability of current cutting-edge COS models on the positive samples of \ourDataset. Regrading the code availability, we here choose four top-performing COS approaches: SINetV2~\cite{fan2022concealed}, DGNet~\cite{ji2023gradient}, CamoFormer-P~\cite{yin2023camoformer}, and HitNet~\cite{hu2023high}. As reported in \tabref{tab:cds_zero_performance}, our observations indicate that these models are not effective in handling cross-domain samples, highlighting the need for further exploration of the domain gap between natural scene and downstream applications.

\section{Conclusion}

This paper aims to provide an overview of deep learning techniques tailored for concealed scene understanding (CSU). To help the readers view the global landscape of this field, we have made four contributions:
Firstly, we provide a detailed survey of CSU, which includes its background, taxonomy, task-specific challenges, and advances in the deep learning era. To the best of our knowledge, this survey is the most comprehensive one to date.
Secondly, we have created the largest and most up-to-date benchmark for concealed object segmentation (COS), which is a foundational and prosperous direction at CSU. This benchmark allows for a quantitative comparison of state-of-the-art techniques.
Thirdly, we have \rev{collected} the largest concealed defect segmentation dataset, CDS2K, by including hard cases from diverse industrial scenarios. We have also constructed a comprehensive benchmark to evaluate the generalizability of deep CSU in practical scenarios.
Finally, we discuss open problems and potential directions for this community. We aim to encourage further research and development in this area.

We would conclude from the following perspectives.
(1) \textbf{Model.}
The most common practice is based on the architecture of sharing UNet, which is enhanced by  various attention modules.
In addition, injecting extra priors and/or introducing auxiliary tasks improve the performance, while there are many potential problems to explore.
(2) \textbf{Training.} 
Fully-supervised learning is the mainstream strategy in COS, but few researchers have addressed the challenge caused by insufficient data or labels. CRNet~\cite{he2023weakly} is a good attempt to alleviate this issue.
(3) \textbf{Dataset.}
The existing datasets are still not large and diverse enough. This community needs more concealed samples involving more domains (\eg, autonomous driving and clinical diagnosis).
(4) \textbf{Performance.}
Transformer and ConvNext based models outperform other competitors by a clear margin. Cost-performance tradeoff is still under-studied, for which  DGNet \cite{ji2023gradient} is a good attempt.
(5) \textbf{Metric.} There is no well-defined metrics that can consider the different camouflage degree of different data to give a comprehensive evaluation. This causes unfair comparisons.

Besides, existing CSU methods focus on the appearance attributes of the concealed scenes (\eg, color, texture, boundary) to distinguish concealed objects without enough perception and output from the semantic perspective (\eg, relationships between objects). 
However, semantics is a good tool for bridging the human and machine intelligence gap.
Therefore, beyond the visual space, semantic level awareness is key to the next-generation concealed visual perception.
In the future, CSU models should incorporate various semantic abilities, including integrating high-level semantics, learning vision-language knowledge~\cite{ji2023masked}, and modeling interactions across objects.

We hope that this survey provides a detailed overview for new researchers, presents a convenient reference for relevant experts, and encourages future research.

\ifCLASSOPTIONcaptionsoff
  \newpage
\fi

{
\bibliographystyle{IEEEtran}
\bibliography{bibliography}
}

\vfill

\end{document}